\pdfoutput=1
\documentclass{article}

\usepackage{iclr2027_conference,times}

\usepackage[hidelinks]{hyperref}
\usepackage{url}
\usepackage{latexsym}
\usepackage[T1]{fontenc}

\usepackage[utf8]{inputenc}
\usepackage{microtype}
\usepackage{graphicx}
\usepackage{amsmath}
\usepackage{amssymb}
\usepackage{booktabs}
\usepackage{tikz}
\usetikzlibrary{arrows.meta,positioning,calc,plotmarks}
\usepackage{enumitem}
\usepackage{fancyvrb}
\usepackage{placeins}
\usepackage{float}
\usepackage{longtable}
\usepackage{etoolbox}

\usepackage{wrapfig}
\usepackage{enumitem}
\setlist[itemize]{leftmargin=*}

\AddToHook{env/table/begin}{\setlength{\abovecaptionskip}{1.25pt}\setlength{\belowcaptionskip}{4pt}}
\AddToHook{env/table*/begin}{\setlength{\abovecaptionskip}{1.25pt}\setlength{\belowcaptionskip}{4pt}}
\makeatletter
\patchcmd{\LT@makecaption}{\vskip\baselineskip}{\vskip5pt}{}{\PackageError{manuscript}{Longtable caption spacing was not applied}{Check the longtable caption definition.}}
\makeatother

\newcommand{\best}[1]{\textbf{#1}}
\definecolor{llmBlue}{HTML}{DDF0FA}
\definecolor{llmGreen}{HTML}{E8F8E8}
\definecolor{llmOrange}{HTML}{FFF0D9}
\definecolor{llmRed}{HTML}{FDE6E6}
\definecolor{llmGray}{HTML}{F3F3F3}
\DefineVerbatimEnvironment{PromptBlock}{Verbatim}{
  fontsize=\footnotesize,
  frame=single,
  framerule=0.3pt,
  framesep=3pt,
  rulecolor=\color{black!35}
}

\title{AG-CoT: Verified Algorithmic Traces for\\LLM Program Synthesis on Clifford Circuits}

\iclrfinalcopy
\author{Lu Wei$^{1}$\quad Yufeng Wang$^{2}$\quad Chenfeng Cao$^{3}$\quad Lu Pang$^{2}$\quad Haibin Ling$^{4}$ \\
$^{1}$Data Science Department, $^{2}$Department of Computer Science, Stony Brook University \\
$^{3}$Dahlem Center for Complex Quantum Systems, Freie Universit\"at Berlin\quad $^{4}$Westlake University \\
\texttt{luw744895@gmail.com}, \texttt{yufeng.wang.2@stonybrook.edu}, \\
\texttt{chenfeng.cao@fu-berlin.de}, \texttt{lu.pang@stonybrook.edu}, \\
\texttt{linghaibin@westlake.edu.cn}
}

\begin{document}
\raggedbottom
\maketitle
\lhead{Preprint. Under review.}
\suppressfloats[t]

\begin{abstract}
Scientific code generation can produce executable programs that fail to compute the intended scientific object.
We study this problem in language-model synthesis of Clifford circuits, which prepare the stabilizer states used in quantum error correction and admit exact classical verification.
In our target-conditioned framework, each target is given as compact signed stabilizer generators, and an exact verifier checks the generated OpenQASM circuits.
We supervise models with Aaronson--Gottesman chain-of-thought (AG-CoT) traces checked by the verifier, and continue training on model generations that the verifier accepts.
Across two independently trained model families (3B and 7B), AG-CoT supervision multiplies greedy-decode state-equivalence accuracy by four to six times over circuit-only baselines, and verifier-filtered continuation training adds a further consistent gain atop both.
A complementary 32B study shows that supervised models achieve near-perfect syntax and Clifford validity while the strongest direct model reaches 6.14\% state equivalence per target, rising to over 10\% under verifier-guided selection with multiple candidates.
These results show that algorithmic trace supervision gives a large, statistically significant gain in both model families and that verifier-filtered continuation adds a further repeated gain.
The persistent gap between Clifford validity and state equivalence confirms that exact verification is necessary: a circuit can be syntactically and physically valid yet prepare the wrong quantum state.
\end{abstract}

\vspace{-4pt}
\section{Introduction}
\vspace{-4pt}

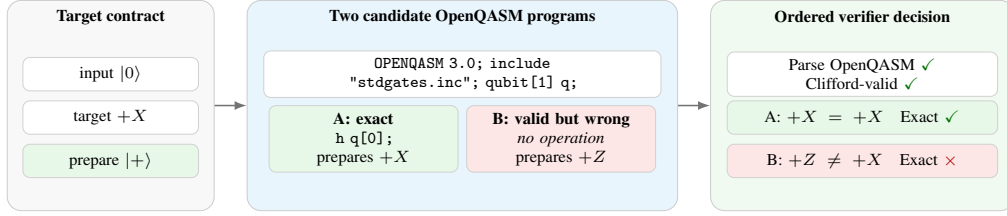
\begin{figure}[!t]
\vspace{1pt}
\centering
\resizebox{0.96\textwidth}{!}{%
\begin{tikzpicture}[
    font=\scriptsize,
    card/.style={draw=black!20, rounded corners=6pt, fill=white, align=center, inner sep=5pt, minimum height=3.15cm},
    title/.style={font=\bfseries\scriptsize, align=center},
    chip/.style={draw=black!18, rounded corners=3pt, fill=white, align=center, inner sep=3pt, minimum height=0.50cm},
    flow/.style={-{Latex[length=1.8mm]}, line width=0.65pt, draw=black!55}
]
\node[card, fill=llmGray!60, minimum width=3.10cm] (target) at (0,0) {};
\node[title] at ($(target.north)+(0,-0.25)$) {Target contract};
\node[chip, text width=2.42cm] at ($(target.center)+(0,0.47)$) {input $|0\rangle$};
\node[chip, text width=2.42cm] at ($(target.center)+(0,-0.18)$) {target $+X$};
\node[chip, text width=2.42cm, fill=llmGreen] at ($(target.center)+(0,-0.83)$) {prepare $|+\rangle$};

\node[card, fill=llmBlue!65, minimum width=6.45cm, right=0.48cm of target] (programs) {};
\node[title] at ($(programs.north)+(0,-0.25)$) {Two candidate OpenQASM programs};
\node[chip, text width=5.74cm] at ($(programs.center)+(0,0.47)$)
{{\ttfamily OPENQASM 3.0; include "stdgates.inc"; qubit[1] q;}};
\node[chip, text width=2.62cm, fill=llmGreen] at ($(programs.center)+(-1.48,-0.50)$)
{\textbf{A: exact}\\{\ttfamily h q[0];}\\prepares $+X$};
\node[chip, text width=2.62cm, fill=llmRed] at ($(programs.center)+(1.48,-0.50)$)
{\textbf{B: valid but wrong}\\\textit{no operation}\\prepares $+Z$};

\node[card, fill=llmGreen!65, minimum width=4.55cm, right=0.48cm of programs] (decision) {};
\node[title] at ($(decision.north)+(0,-0.25)$) {Ordered verifier decision};
\node[chip, text width=3.84cm] at ($(decision.center)+(0,0.47)$)
{Parse OpenQASM \textcolor{green!45!black}{\checkmark}\\Clifford-valid \textcolor{green!45!black}{\checkmark}};
\node[chip, text width=3.84cm, fill=llmGreen] at ($(decision.center)+(0,-0.18)$)
{A: $+X = +X$\quad Exact \textcolor{green!45!black}{\checkmark}};
\node[chip, text width=3.84cm, fill=llmRed] at ($(decision.center)+(0,-0.83)$)
{B: $+Z \neq +X$\quad Exact \textcolor{red!70!black}{$\times$}};

\draw[flow] (target.east) -- (programs.west);
\draw[flow] (programs.east) -- (decision.west);
\end{tikzpicture}
}
\caption{Running verifier example. Both candidates parse and remain Clifford-valid, but Candidate B is rejected because it prepares $+Z$ rather than the requested $+X$ target.}
\label{fig:running-example}
\par\vspace{2pt}
\end{figure}

Language models now generate executable code, and execution-based benchmarks made behavioral correctness the standard for judging it \citep{chen2021evaluatinglargelanguagemodels,austin2021programsynthesislargelanguage,hendrycks2021measuringcodingchallengecompetence}.
Recent work extends this standard with executable environments, semantic judgment, and equivalence checking \citep{wang-etal-2023-execution,huang-etal-2024-da,yan-etal-2024-codescope,tong-zhang-2024-codejudge,wei-etal-2025-equibench}.
Quantum programs have become an active target for language models, with fine-tuned generators, agentic reinforcement learning and dedicated benchmarks \citep{jern2025agentqfinetuninglargelanguage,yu2025quasarquantumassemblycode,yu2026vista,yang2025qcircuitbenchlargescaledatasetbenchmarking,guo2025quanbench}.
They also sharpen the gap between running and being correct, since a circuit can parse and execute yet prepare the wrong state.
In the running example of Figure~\ref{fig:running-example}, two one-qubit programs are both valid, but only one prepares the requested state.

We study this gap in Clifford circuits, the quantum circuits built from Hadamard, phase, and CNOT gates.
Adding the non-Clifford $T$ gate to these gates yields a universal gate set for quantum computation \citep{boykin1999universalfaulttolerant}.
Clifford circuits prepare the stabilizer states that encode quantum error-correcting codes, serve as resource states for measurement-based quantum computation, and model large-scale quantum dynamics \citep{gottesman1997stabilizercodesquantumerror,raussendorf2001onewayquantumcomputer,raussendorf2003measurementbasedquantumcomputation,li2019measurementdrivenentanglementtransition}.
Their synthesis and optimization remain active compilation problems \citep{bravyi2021clifford,schneider2023sat,paz2026stabilizerbench}.
Clifford circuits are exactly checkable: an $n$-qubit stabilizer state is fixed by $n$ signed Pauli generators, and a Clifford circuit can be simulated efficiently on a classical computer instead of tracking $2^n$ amplitudes \citep{aaronson2008improvedsimulationstabilizercircuits}.
Clifford state preparation therefore gives every output an exact and inexpensive correctness check, which most code generation lacks, and the Aaronson--Gottesman algorithm provides a step-by-step derivation of a correct circuit for every target \citep{aaronson2008improvedsimulationstabilizercircuits}.
These two properties make Clifford state preparation a controlled setting for asking whether supervising the derivation and training on verified outputs improve the semantic correctness of generated programs, a question that applies to LLM program synthesis in general.

In the recent benchmark, QCircuitBench, every evaluated model, including GPT-4o, scores 0.0000 fidelity, the squared overlap between the prepared and target states, on Clifford state preparation under both 1-shot and 5-shot prompting \citep{yang2025qcircuitbenchlargescaledatasetbenchmarking}.
We therefore use the exact check as a training signal rather than only as a score.
First, we supervise models with Aaronson--Gottesman chain-of-thought (AG-CoT) traces, deterministic tableau-reduction steps whose inversion yields the emitted circuit, kept only when the verifier accepts that circuit \citep{aaronson2008improvedsimulationstabilizercircuits}.
Second, we continue supervised training on the model's own outputs that the verifier accepts, a verifier-filtered form of rejection-sampling fine-tuning (RFT) \citep{yuan2023scalingrelationship}.
Each target reaches the model as compact signed stabilizer generators rather than a $2^n$-entry state vector, which shortens a 12-qubit prompt from roughly 16K to roughly 800 characters (Appendix Figure~\ref{fig:target-representation}).
Figure~\ref{fig:training-overview} summarizes the pipeline and results.

Controlled comparisons on the same 3,661 held-out targets, with one greedy output per target, isolate both effects.
For Qwen2.5-3B-Instruct, circuit-only SFT solves 33 targets, AG-CoT SFT solves 201, and three RFT versions solve 213, 215, and 210.
Mistral-7B-Instruct-v0.3 follows the same pattern, from 71 to 316 and then 322.
In a complementary 32B study, supervised models reach approximately 99.9\% syntax and Clifford validity, yet the strongest direct model prepares the target state for only 6.14\% of prompts.

To summarize, we make four main contributions:
\begin{itemize}
    \item We supervise circuit synthesis with verifier-checked AG-CoT traces, which pair each emitted circuit with the Aaronson--Gottesman reduction that produced it.
Against a matched circuit-only control, this supervision raises single-output task success from 0.90\% to 5.49\% in Qwen and from 1.94\% to 8.63\% in Mistral.
    \item We continue training on verifier-accepted model outputs, which adds a smaller observed increase in all three Qwen versions (5.74\% to 5.87\%) and in Mistral (8.80\%).
    \item Building on QCircuitBench instances and reference circuits \citep{yang2025qcircuitbenchlargescaledatasetbenchmarking}, we state each target as signed stabilizer generators and keep OpenQASM generation with exact target-state semantics, so one verifier certifies supervision, selects training data, and scores outputs.
    \item We report syntax validity, Clifford validity, saved-label matching, state equivalence, and finite-budget search coverage separately, which shows where valid quantum code fails to become correct state preparation.
\end{itemize}

\vspace{-4pt}
\section{Related Work}
\vspace{-4pt}

\noindent\textbf{Semantic code evaluation.}
Execution-based benchmarks shifted code evaluation from textual overlap to behavioral correctness \citep{chen2021evaluatinglargelanguagemodels,austin2021programsynthesislargelanguage,hendrycks2021measuringcodingchallengecompetence}.
Recent benchmarks and evaluators extend this direction to executable environments, semantic judgment, and equivalence checking \citep{wang-etal-2023-execution,yan-etal-2024-codescope,huang-etal-2024-da,tong-zhang-2024-codejudge,wei-etal-2025-equibench}.
We share this semantic motivation and use an exact quantum-domain oracle for training as well as evaluation, where correctness means preparing the target stabilizer state rather than passing sampled tests or matching surface code.

\noindent\textbf{Quantum circuit generation and synthesis.}
Classical and generative circuit synthesis includes template and SAT methods for Clifford circuits and text-conditioned diffusion models for quantum operations \citep{bravyi2021clifford,schneider2023sat,furrutter2024quantum}.
Quantum-code benchmarks evaluate executable and semantic behavior through state preparation, process fidelity, and simulator feedback \citep{yang2025qcircuitbenchlargescaledatasetbenchmarking,guo2025quanbench,mikuriya2025qcoder,paz2026stabilizerbench}.
Language-model systems span task-specific fine-tuning, tool-augmented rewards, verifier-in-the-loop reinforcement learning, operator-conditioned synthesis, and evaluator-guided test-time search \citep{jern2025agentqfinetuninglargelanguage,yu2025quasarquantumassemblycode,yu2026vista,feris2026aligningquantumoperators,macaronepalmieri2026quantum}.
Meta-design instead generates programs that construct families of quantum experiments and circuits \citep{arlt2024meta}.
We study Clifford state preparation from signed stabilizer targets.
This representation permits exact verification of the prepared state after parsing and Clifford validity checks.

\noindent\textbf{Verifier-guided learning and search.}
Verifiers can guide training examples, reward assignment, and inference-time candidate selection.
Chain-of-thought traces expose intermediate reasoning steps \citep{wei2022chain}.
Rejection sampling fine-tuning retrains a model on filtered successful generations \citep{yuan2023scalingrelationship}, while GRPO forms group-relative policy updates from sampled rewards \citep{shao2024deepseekmath}.
Reward-side analyses, Best-of-$N$ or test-time scaling show that training reward and Pass@$N$ behavior need not equal single-sample reliability \citep{li2022competition,brown2024largelanguagemonkeysscaling,he-etal-2025-rewarding,li-etal-2025-test}.
Building on QCircuitBench targets and the stabilizer reduction of Aaronson and Gottesman, we study circuit synthesis conditioned on exact signed stabilizer targets \citep{yang2025qcircuitbenchlargescaledatasetbenchmarking,aaronson2008improvedsimulationstabilizercircuits}.
Our training design pairs deterministic traces aligned with emitted circuits with a matched circuit only control, then continues supervised training on model outputs selected by the verifier.
The same verifier checks training completions and generated circuits.

\vspace{-4pt}
\section{Task and Verifier}
\vspace{-4pt}
\label{sec:task-verifier}

Our task is to write a quantum circuit that prepares a requested quantum state.
The model reads a description of the target state and emits a program, which counts as correct only if it prepares exactly that state.
For a target instance $x$ on $n$ qubits, the prompt $p_x$ specifies the state through signed Pauli generators, not a target unitary or reference circuit.
The output $y$ is an OpenQASM program that must prepare this state from $|0\rangle^{\otimes n}$ \citep{cross2022openqasm}.
Our Qiskit pipeline parses the program, checks Clifford validity, and compares the prepared state with the target \citep{javadiabhari2024qiskit}.
Appendix Table~\ref{tab:appendix-formal-task-boundary} gives the full protocol.

\noindent\textbf{Stabilizer-state tableau semantics.}
The verifier is possible because Clifford state preparation has an exact finite stabilizer representation.
A stabilizer state $|\psi\rangle$ can be specified by $n$ independent commuting signed Pauli generators $S_1,\ldots,S_n$ such that $S_i|\psi\rangle=|\psi\rangle$ for every generator index $i$.
The target tableau is therefore $T_x=(M_x,r_x)$, where $M_x\in\{0,1\}^{n\times 2n}$ stores the binary $X/Z$ support of the target stabilizer generators and $r_x\in\{0,1\}^{n}$ stores their signs \citep{aaronson2008improvedsimulationstabilizercircuits}.
Here $X$ and $Z$ are Pauli operators, and the implementation stores the generators as signed Pauli strings.
Different generator bases can describe the same state, so label equality and state equivalence are distinct checks.
Appendix~\ref{sec:appendix-formal-task-verifier} gives the signed Pauli expansion and gate update equations.

Clifford gates map signed Pauli operators to signed Pauli operators under conjugation, so a stabilizer state remains within the same compact representation throughout the circuit.
Each gate updates the binary support and sign bits exactly, allowing the verifier to compare the prepared state with the target without reconstructing a numerical state vector \citep{aaronson2008improvedsimulationstabilizercircuits}.
This structure makes exact target checks practical for supervision, reward assignment, and candidate selection.
An extension to generic state preparation would require a representation for non-stabilizer states and a state-fidelity threshold fixed before evaluation.
An extension to full unitary synthesis would instead compare the candidate and target unitaries up to global phase, with an explicit numerical tolerance for approximate evaluation.

\noindent\textbf{Why the prompt uses tableaus.}
The tableau provides a compact exact text interface to the target state.
Because prompting alone scores zero on the QCircuitBench Clifford targets (Section~1), we test whether explicit signed-generator targets, together with training, improve state preparation.
A general pure-state vector contains $2^n$ complex amplitudes, whereas the stabilizer tableau stores $2n^2+n$ bits before formatting.
In the same 12-qubit conversion example, 4,096 complex amplitudes occupy approximately 16,000 printed characters, compared with approximately 800 for the tableau prompt.
Appendix~\ref{sec:appendix-dataset-preprocessing} gives the full representation and conversion protocol.

\noindent\textbf{Verifier as a semantic decision chain.}
The verifier records the first failed stage or  terminal target match.
Let $P(y)$ be 1 if $y$ parses as OpenQASM and 0 otherwise.
Let $K(y)$ be 1 if the parsed circuit can be converted to a Qiskit Clifford object and 0 otherwise. If $P(y)=0$, then $K(y)=0$.
For a parsed Clifford circuit, let $T_y$ be the tableau prepared by executing it from $|0\rangle^{\otimes n}$, and let $\mathcal{L}(T)$ denote the Qiskit signed stabilizer-label set of a tableau $T$.
The strict decision $\mathrm{Exact}(x,y)$ requires $P(y)\!\!=\!\!K(y)\!\!=\!\!1$ and $\mathcal{L}(T_y)\!=\!\mathcal{L}(T_x)$, ignoring generator order but not its basis.
It is stricter than syntax validity, qubit-count correctness, expectation-value proximity or sampled distributional similarity.
It checks state preparation under the saved-label convention, not full Clifford-unitary equivalence.
Let $\mathrm{Equiv}(T_y,T_x)$ be 1 when the stabilizer states represented by $T_y$ and $T_x$ are equivalent under Qiskit's \texttt{StabilizerState.equiv} routine, and 0 otherwise.
Let $\mathbf{1}[\cdot]$ be the indicator function.
The corresponding direct semantic metric is $\mathrm{StateEq}(x,y)\!=\!\mathbf{1}[P(y)\!=\!1 \land K(y)\!=\!1 \land \mathrm{Equiv}(T_y,T_x)=1]$.
StateEq@1 hence removes generator-basis sensitivity while preserving the same parsing, Clifford-conversion and state-preparation boundary as Exact@1$_{\mathrm{lab.}}$.
Reporting both metrics separates a representation-convention penalty from genuine target-state failure.

\noindent\textbf{Verifier stages and metric interpretation.}
Syntax validity reports whether $P(y)=1$.
Clifford validity reports whether $P(y)=K(y)=1$.
Exact@1 reports whether one generated candidate satisfies $\mathrm{Exact}(x,y)=1$.
For a candidate budget $N$, Pass@$N$ measures the probability of at least one strict success, averaged over targets.
StateEq-Pass@$N$ instead uses state equivalence as each candidate's success criterion.
The search runs sample 64 candidates per target under one decoding configuration.
Pass@64 is observed target coverage, while smaller-budget values are estimates from the same pool using the estimator of \citet{chen2021evaluatinglargelanguagemodels}, given in Appendix~\ref{sec:appendix-formal-task-verifier}.
Estimated Pass@1 averages the fraction of correct candidates in that pool and is distinct from direct Exact@1 measured in the separate one-output runs.

\vspace{-4pt}
\section{Target-Conditioned Supervision and Verifier-Filtered Training}
\vspace{-4pt}

\begin{figure}[!t]
\vspace{-3.25pt}
\centering
\includegraphics[width=.97\linewidth]{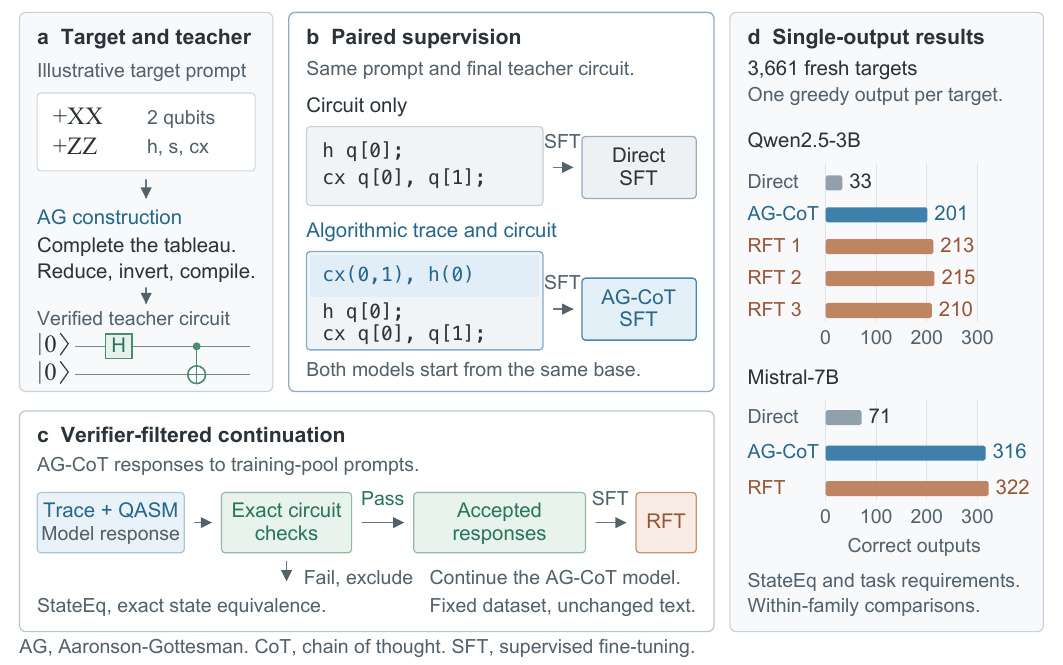}
\vspace{-.5pt}
\caption{Target-conditioned circuit synthesis with algorithmic trace supervision and verifier-filtered continuation. Panel \textbf{a} constructs a verified teacher circuit from signed target generators; \textbf{b} pairs circuit-only and AG-CoT supervision with the same prompt and final circuit. The two-qubit example is schematic, with the blue reduction operations inverted to obtain the preparation circuit. Panel \textbf{c} forms a fixed dataset of accepted AG-CoT responses for supervised continuation (RFT). Acceptance checks the emitted circuit and task requirements. Panel \textbf{d} shows correct single outputs on the same 3,661 test targets for each trained system in Table~\ref{tab:3b-training-comparison}. The three Qwen RFT versions remain separate.}
\label{fig:training-overview}
\end{figure}

\noindent\textbf{Method overview.}
Our training pipeline uses the exact target representation and verifier from Section~\ref{sec:task-verifier} to construct supervised completions and select successful model generations.
QCircuitBench supplies the Clifford instances and reference circuits for the 32B study \citep{yang2025qcircuitbenchlargescaledatasetbenchmarking}.
The pipeline has two components, AG-CoT supervision and verifier-filtered continuation.
AG-CoT completions contain the Aaronson--Gottesman reduction steps from which each circuit is constructed, so the supervision covers the synthesis procedure as well as its output.
Direct SFT omits these steps and serves as the control for trace supervision.
The target-conditioned construction used for Qwen and Mistral builds its teacher directly from signed target generators, whereas the 32B construction derives traces from reference Clifford circuits.
Figure~\ref{fig:training-overview} summarizes the paired supervision and filtered continuation used in the controlled Qwen and Mistral comparisons.

\noindent\textbf{Reference-circuit AG-CoT construction.}
AG-CoT constructs supervised completions that expose the synthesis procedure, rather than serving only as an inference-time prompt.
For each instance $x$, we parse its QCircuitBench reference circuit into a Clifford object and apply deterministic Aaronson-Gottesman row reduction \citep{aaronson2008improvedsimulationstabilizercircuits}.
For multi-qubit instances, the trace $\rho_x$ records the circuit $R_x$ that reduces this reference Clifford to the identity, including final phase corrections.
The preparation circuit is $G_x=R_x^{-1}$, the reduction in reverse order with every gate inverted.
Basic AG-CoT groups the reduction operations by qubit and records the signed destabilizer row at the start of each qubit's reduction.
Step AG-CoT additionally records that qubit's destabilizer row after each operation in its reduction.
Both variants list any final phase-correction operations separately.
The completion $c_x$ places $\rho_x$ inside a \texttt{<think>} block, followed by the OpenQASM program $y_x$ for that preparation circuit.
The corpus keeps only verified pairs, $\mathcal{D}_{\mathrm{AG}}^{32\mathrm{B}}=\{(p_x,c_x):\mathrm{Exact}(x,y_x)=1\}$, under the strict saved-label decision.
Appendix~\ref{sec:ag-cot-appendix} gives the full acceptance notation, Basic and Step recording rules, and the single-qubit direct-synthesis branch.
The construction yields 38,940 accepted examples with no generation failures. 
Appendix Figure~\ref{fig:verifier-pipeline} follows the 32B study from its target representation through supervision and evaluation.

\begin{figure}[!t]
\centering
\includegraphics[width=.97\linewidth]{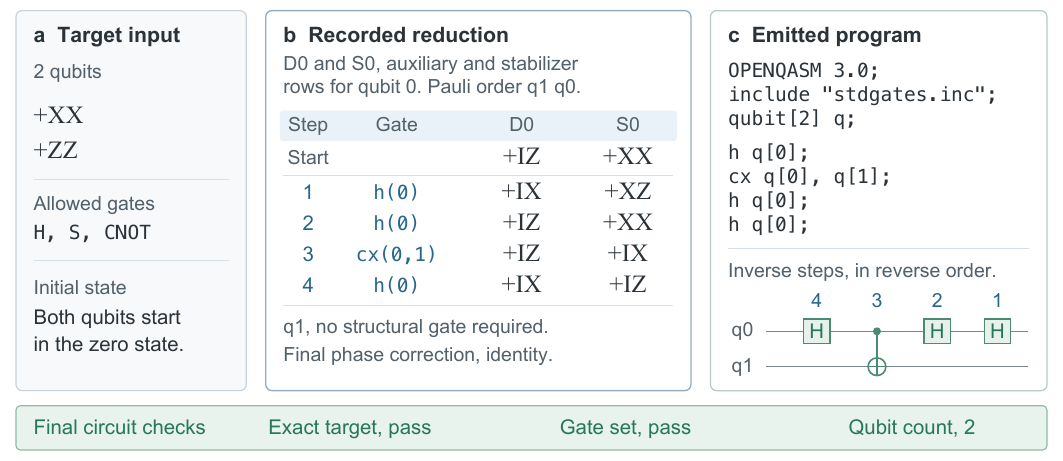}
\vspace{-.75pt}
\caption{%
A target-conditioned AG-CoT training example. The reformatted input and trace excerpt retain all four reduction operations and the corresponding auxiliary and stabilizer rows for qubit 0. Inverting the reduction gives the complete program shown on the right. Circuit labels identify the reduction steps in reverse order. The adjacent final Hadamard gates remain as emitted by the teacher.}
\label{fig:ag-cot-completion-main}
\end{figure}

\noindent\textbf{Target-conditioned AG-CoT construction.}
For the Qwen2.5-3B-Instruct and matched Mistral comparisons, a separate construction starts from the signed stabilizer generators in the prompt rather than from the reference circuit.
It completes those state generators to one full Clifford tableau, applies Aaronson-Gottesman reduction, inverts the reduction, compiles the resulting preparation circuit to \texttt{\{h, s, cx\}}, and verifies the emitted program with StateEq and the task requirements used in these comparisons.
The source QASM is used only to confirm that it prepares the recorded target state.
The Mistral comparison uses the same target-conditioned teacher construction and paired circuit-only supervision.
Figure~\ref{fig:ag-cot-completion-main} expands one recorded target-conditioned training example, and Appendix~\ref{sec:ag-cot-appendix} defines the auxiliary destabilizer rows and the full construction protocol.
In both constructions, the teacher trace and final circuit come from the same deterministic procedure, and filtering verifies the circuit against the target.
Generated-output evaluation instead executes the emitted QASM without checking every intermediate model-generated row, so final-circuit mismatch is not a separately measured trace-row error rate.

\noindent\textbf{Paired supervision and training objective.}
The Qwen2.5-3B-Instruct comparison includes a circuit-only supervised baseline, Direct SFT.
Its completion is the target-conditioned 3B AG-CoT completion $c_x$ with the \texttt{<think>} block removed, leaving the final QASM code fence $y_x$ verbatim.
All other fields, including the target-conditioned prompt and target identifier, remain unchanged.
The paired records thus share the same input and teacher circuit, while AG-CoT additionally supervises the algorithmic trace.
Basic AG-CoT SFT uses compact verified traces with one trace line per qubit.
Step AG-CoT SFT exposes intermediate destabilizer-row states after each reduction operation, testing whether denser algorithmic supervision changes exact target match.
Let $r$ identify the reference-circuit or target-conditioned construction, $\mathcal{D}_{\mathrm{AG}}^{(r)}$ its accepted prompt-completion pairs, and $\pi_\theta$ the language model with parameters $\theta$.
Each AG-CoT SFT model minimizes the negative conditional log likelihood $\mathcal{L}_{\mathrm{SFT}}$ of its completions, $\mathcal{L}_{\mathrm{SFT}}(\theta)=-\sum_{(p_x,c_x)\in \mathcal{D}_{\mathrm{AG}}^{(r)}}\log \pi_\theta(c_x\mid p_x)$.

\noindent\textbf{Verifier-guided training and candidate selection.}
Beyond the paired SFT comparison, the verifier supports reward-based training, filtered continuation, and test-time candidate selection.
We evaluate GRPO by held-out Exact@1, using the partial saved-label reward in Appendix~\ref{sec:appendix-training-search} only as the training signal.
Verifier-filtered RFT samples candidates, keeps only verifier-confirmed successes, and fine-tunes on those completions \citep{yuan2023scalingrelationship}.
Best-of-64 does not change the model. It samples 64 candidates per target and measures whether verifier selection finds at least one candidate that passes the specified target criterion.

\noindent\textbf{Verifier-filtered supervised continuation.}
For Qwen2.5-3B-Instruct, RFT further adapts the AG-CoT model on its own verifier-selected completions.
We sample one response per target prompt in the training pool.
The fixed dataset is $\mathcal{D}_{\mathrm{RFT}}^{3\mathrm{B}}=\{(p_x,y):\mathrm{Acc}(x,y)=1\}$, where $\mathrm{Acc}(x,y)=1$ when generation completes normally, $y$ contains one \texttt{<think>} block followed by one complete fenced QASM program, and that program begins with \texttt{OPENQASM 3.0;}, parses, is Clifford-valid, uses only \texttt{\{h, s, cx\}}, has the requested qubit count, and passes StateEq.
Strict saved-label equality is recorded but not required.
Each accepted response is kept verbatim with its original prompt, and RFT continues supervised training from the AG-CoT adapter on this fixed dataset with prompt tokens masked from the loss.

\vspace{-4pt}
\section{Experimental Setup}
\vspace{-4pt}
\label{sec:experimental-setup}

The experiments use Clifford state-preparation instances from the QCircuitBench random-circuit synthesis setting, where exact stabilizer-tableau verification is available.
Appendix~\ref{sec:prompt-decoding-record} records the prompt and decoding settings for the reported evaluations.

\noindent\textbf{Controlled Qwen and Mistral comparison.}
The test set contains 3,661 mutually distinct targets, generated after all six Qwen checkpoints were frozen. None shares its canonical identity (the row-reduced signed symplectic representation of its full stabilizer group) with the 37,071 training and development target identities.
On this fixed test set, we compare the Qwen2.5-3B-Instruct base model \citep{yang2024qwen25}, Direct SFT, AG-CoT SFT, and three verifier-filtered RFT versions.
All models receive the same prompts and generate one greedy completion per target at temperature zero, without retries.
The 8,192-token context permits up to 8,000 new tokens, with the output cap reduced to fit each prompt.
QASM is extracted by a fixed rule without repair or format conversion.
Task success requires StateEq, compliance with the requested \texttt{\{h, s, cx\}} gate set, the required \texttt{OPENQASM 3.0;} header, and an output below the per-request token limit.
Invalid and truncated outputs fail.
The matched Mistral-7B-Instruct-v0.3 comparison \citep{jiang2023mistral,mistralai2024mistral7binstructv03} uses the same test targets and task-success rule, with its training and evaluation settings specified in Appendix~\ref{app:mistral-training}.

\noindent\textbf{Complementary 32B study.}
The 32B study uses a split of 35,046 AG-CoT training examples and 3,894 held-out test prompts, on which all of its direct results are measured.
It compares five direct-generation systems and one search-time variant, all built on Qwen2.5-Coder-32B-Instruct \citep{hui2024qwen25coder}.
Zero-shot uses this base model directly.
Basic and Step AG-CoT SFT use the per-qubit and per-operation trace formats defined above.
GRPO uses the partial saved-label reward defined in Appendix~\ref{sec:appendix-training-search}.
Verifier-filtered RFT samples candidates, filters exact successes with the verifier, and fine-tunes on verified completions.
Best-of-64 is reported separately because it uses repeated sampling and verifier selection rather than a single circuit.

\noindent\textbf{Metrics.}
We report syntax validity, Clifford validity, Exact@1$_{\mathrm{lab.}}$, StateEq@1, Pass@64, and StateEq-Pass@64.
Section~\ref{sec:task-verifier} defines these metrics and their distinct terminal criteria.
The controlled comparisons report StateEq-based task success under the additional requirements above.

\section{Results and Analysis}
\label{sec:results}

\begin{wraptable}{r}{.39\linewidth}
\centering
\caption{Single-output training comparison on the same 3,661 target prompts. Each row uses one greedy completion per target and no program repair. %
}
\label{tab:3b-training-comparison}
\resizebox{.92\linewidth}{!}{ 
\begin{tabular}%
{@{\hspace{.51mm}}l@{\hspace{.51mm}} @{\hspace{.51mm}}r@{\hspace{1.51mm}} @{\hspace{1.51mm}}c@{\hspace{.51mm}} }
\toprule
System & Correct/total & Suc(\%) \\
\midrule
\multicolumn{3}{l}{\textit{Qwen2.5-3B-Instruct}} \\
Base model & 0 / 3,661 & 0.00 \\
Direct SFT & 33 / 3,661 & 0.90 \\
AG-CoT SFT & 201 / 3,661 & 5.49 \\
RFT ver. 1 & 213 / 3,661 & 5.82 \\
RFT ver. 2 & 215 / 3,661 & \best{5.87} \\
RFT ver. 3 & 210 / 3,661 & 5.74 \\
\midrule
\multicolumn{3}{l}{\textit{Mistral-7B-Instruct-v0.3}} \\
Direct SFT & 71 / 3,661 & 1.94 \\
AG-CoT SFT & 316 / 3,661 & 8.63 \\
RFT & 322 / 3,661 & \best{8.80} \\
\bottomrule
\end{tabular}
}
\vspace{-3mm}
\end{wraptable}

We first report the controlled Qwen and Mistral comparisons, then the complementary 32B study of training variants, metric choice and verifier-selected search.
The separate development comparison is reported in Appendix Table~\ref{tab:cross-family-direct-sft}.
Because the model family, model size, target set, terminal decision, and candidate budget differ across these protocols, their rows are not treated as a model-size ranking or as one continuous training curve.
Within each protocol, the comparisons test the effect of algorithmic supervision, verifier-filtered continuation, or verifier-selected search.

\subsection{Controlled Training Comparisons}
\label{sec:3b-training-results}

Table~\ref{tab:3b-training-comparison} compares the training methods within Qwen2.5-3B-Instruct and Mistral-7B-Instruct-v0.3 under the common single-output protocol (Appendix~\ref{sec:prompt-decoding-record}).
The three Qwen RFT versions are evaluated separately. %
In these evaluations, every StateEq success also meets the gate, header, and output-length requirements, so the task-success counts equal the StateEq counts.

\begin{wrapfigure}{r}{0.39\textwidth}
\vspace{-31pt}
\centering
\centering
\setlength{\abovecaptionskip}{-4.2pt}
\setlength{\belowcaptionskip}{0pt}
\begin{tikzpicture}[x=0.45cm,y=0.30cm,font=\fontsize{8}{9.5}\selectfont]
\node[font=\bfseries\fontsize{8}{9.5}\selectfont] at (4.5,12.8) {Qwen2.5-3B-Instruct};
\draw[black!65,dashed,line width=0.7pt] (0,11.5) -- (0.8,11.5);
\draw[black!65,fill=white] (0.4,11.5) circle (1.4pt);
\node[anchor=west,inner sep=2pt] at (0.8,11.5) {Direct SFT};
\draw[blue!65!black,line width=0.9pt] (5,11.5) -- (5.8,11.5);
\fill[blue!65!black] (5.4,11.5) +(-1.4pt,-1.4pt) rectangle +(1.4pt,1.4pt);
\node[anchor=west,inner sep=2pt] at (5.8,11.5) {AG-CoT SFT};
\foreach \rate in {0,2,4,6,8,10} {
    \draw[black!12] (0,\rate) -- (9,\rate);
    \node[anchor=east,inner sep=3pt] at (0,\rate) {\rate};
}
\draw[black!60] (0,10.5) -- (0,0) -- (9.3,0);
\foreach \qubits in {3,...,12} {
    \draw[black!60] ({\qubits-3},0) -- ++(0,-0.15);
    \node[anchor=north,inner sep=3pt] at ({\qubits-3},-0.15) {\qubits};
}
\node[rotate=90] at (-1.35,5.25) {Task success (\%)};
\node at (4.5,-2.2) {Target qubits};
\draw[black!65,dashed,line width=0.7pt]
    plot[mark=o,mark size=1.8pt,mark options={solid,draw=black!65,fill=white}] coordinates {
    (0,{100*0/30}) (1,{100*1/95}) (2,{100*3/135})
    (3,{100*2/201}) (4,{100*4/282}) (5,{100*7/355})
    (6,{100*6/454}) (7,{100*4/596}) (8,{100*4/672}) (9,{100*2/841})};
\draw[blue!65!black,line width=0.9pt]
    plot[mark=square*,mark size=1.8pt,mark options={draw=blue!65!black,fill=blue!65!black}] coordinates {
    (0,{100*3/30}) (1,{100*8/95}) (2,{100*12/135})
    (3,{100*10/201}) (4,{100*16/282}) (5,{100*29/355})
    (6,{100*24/454}) (7,{100*27/596}) (8,{100*32/672}) (9,{100*40/841})};
\end{tikzpicture}
\makeatletter\def\@captype{figure}\makeatother
\caption{Qwen2.5-3B-Instruct task success by target size, following requirements in Sec.~\ref{sec:experimental-setup}. Each rate includes all targets at that size and a greedy output per target (Appendix Table~\ref{tab:3b-success-by-qubits}). %
}
\vspace{-2mm}
\label{fig:3b-success-by-qubits}
\end{wrapfigure}
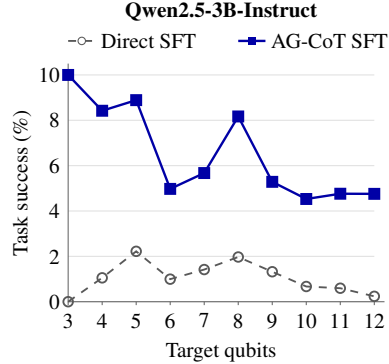

\noindent\textbf{Algorithmic traces improve target-state preparation.}
The Qwen AG-CoT model solves 201 targets, compared with 33 for Direct SFT, an increase of 168 correct outputs or 4.59 percentage points.
Both models already achieve syntax validity above 99\%, so this difference concerns preparing the requested state rather than simply producing parseable code.
The training comparison supports the value of supervising the algorithmic trace alongside the final circuit, beyond circuit-only supervision.
On the same 3,661 targets, AG-CoT SFT turns 170 Direct SFT failures into successes while losing 2 Direct SFT successes, with a two sided exact McNemar $p$ value of $4.97 \times 10^{-48}$, as reported in Table~\ref{tab:paired-task-success}.
Figure~\ref{fig:3b-success-by-qubits} shows a higher observed task success rate for AG-CoT SFT at every tested qubit count from three to twelve, with exact counts in Appendix Table~\ref{tab:3b-success-by-qubits}.
At twelve qubits, Direct SFT solves 2 of 841 targets and AG-CoT solves 40.

\noindent\textbf{Verifier-filtered continuation adds a further gain.}
The three RFT versions solve 210--215 targets each, exceeding AG-CoT SFT by 9--14 correct outputs.
Their mean task-success rate is 5.81\%, 0.32 percentage points above AG-CoT SFT.
The additional gain is smaller than the Direct SFT to AG-CoT difference, but appears in all three versions.
Here the verifier supplies a fixed set of successful model completions for supervised continuation, rather than rewards for online policy optimization.

\begin{wraptable}{r}{.55\linewidth}
\centering
\vspace{-1.5MM}
\caption{Paired Qwen2.5-3B-Instruct task success comparisons on the 3,661 targets. Earlier only and later only are discordant successes in the order shown. Exact $p$ uses the two sided McNemar test. Holm adjusted $p$ covers the three comparisons from AG-CoT SFT to RFT.}
\label{tab:paired-task-success}
\resizebox{.98\linewidth}{!}{ 
\begin{tabular}%
{@{\hspace{.51mm}}c@{\hspace{.51mm}} @{\hspace{.51mm}}c@{\hspace{.51mm}} @{\hspace{.51mm}}c@{\hspace{.51mm}} @{\hspace{.51mm}}c@{\hspace{.51mm}}@{\hspace{.51mm}}c@{\hspace{.51mm}}}
\toprule
 & Earlier & Later  & Exact & Holm \\
Comparison & only &  only &  $p$ &  $p$ \\
\midrule
Direct SFT to AG-CoT SFT & 2 & 170 & 4.97$\times$10$^{-48}$ & -- \\
AG-CoT SFT to RFT ver. 1 & 7 & 19 & 0.029 & 0.058 \\
AG-CoT SFT to RFT ver. 2 & 7 & 21 & 0.013 & 0.038 \\
AG-CoT SFT to RFT ver. 3 & 9 & 18 & 0.122 & 0.122 \\
\bottomrule
\end{tabular}
}
\vspace{-1pt}
\end{wraptable}

\noindent\textbf{Trace supervision also improves Mistral circuit synthesis.}
We repeat the paired training comparison with Mistral-7B-Instruct-v0.3 using the same 35,005 retained examples for Direct SFT and AG-CoT SFT, with identical training settings apart from completion content.
AG-CoT increases exact task success from 71 to 316 targets, a gain of 245 correct outputs or 6.69 percentage points.
Continuation on 983 verifier-selected unedited model completions reaches 322 successes, a further gain of 6 targets or 0.16 percentage points.
Thus the large gain from trace supervision appears in a second model family, with a smaller observed increase after filtered continuation.
Appendix~\ref{app:mistral-training} specifies the training, filtering and evaluation settings.

\noindent\textbf{Paired outcomes distinguish the large trace gain from the smaller continuation gains.}
Table~\ref{tab:paired-task-success} compares successes on the same Qwen targets, rather than comparing aggregate rates alone.
Across the three RFT versions, continuation gains 19, 21 and 18 targets and loses 7, 7 and 9 relative to AG-CoT SFT.
After Holm adjustment across these three comparisons, version 2 remains below 0.05.
For Mistral, AG-CoT gains 246 targets and loses 1 relative to Direct SFT, with a two sided exact McNemar $p$ value of $2.19\times10^{-72}$.
Mistral RFT gains 24 and loses 18 relative to AG-CoT, with $p=0.441$.
The paired results support the large trace-supervision gain in both families, while the smaller continuation gains are not uniformly statistically significant.

\subsection{Complementary 32B Results}
\label{sec:32b-results}

In the 32B study, Table~\ref{tab:main-results} reports direct syntax validity, Clifford validity, saved-label Exact@1, StateEq@1 and its confidence intervals on 3,894 prompts.
Appendix Table~\ref{tab:appendix-qasm-only-stress} documents an exploratory QASM-only format stress test.
Its supervised output omits the AG-CoT trace, its generation cap is 2,048 tokens, and 83\% of outputs are marked truncated.
We therefore treat it as a format-and-budget diagnostic rather than a method-ranking row.

\begin{wraptable}{r}{.62\linewidth}
\centering
\caption{32B direct results on 3,894 prompts. Rates are percentages. Exact@1$_{\mathrm{lab.}}$ checks saved-label equality, while StateEq@1 checks state equivalence over the same targets. The final column gives the reported 95\% confidence interval for StateEq@1.}
\label{tab:main-results}
\resizebox{1\linewidth}{!}{ 
\begin{tabular}%
{@{\hspace{.1mm}}l@{\hspace{.1mm}} @{\hspace{.1mm}}c@{\hspace{.51mm}} @{\hspace{.51mm}}c@{\hspace{.51mm}} @{\hspace{.51mm}}c@{\hspace{.51mm}}@{\hspace{.51mm}}c@{\hspace{.51mm}} @{\hspace{.51mm}}c@{\hspace{.591mm}} @{\hspace{.51mm}}c@{\hspace{.1mm}} @{\hspace{.1mm}}c@{\hspace{.1mm}}}
\toprule
& \multicolumn{2}{c}{Surface validity} & \multicolumn{2}{c}{Target match} & \\
\cmidrule(lr){2-3}\cmidrule(lr){4-5}
System & Syntax & Clifford & Exact@1$_{\mathrm{lab.}}$ & StateEq@1 & 95\% CI \\
\midrule
Zero-shot & 39.9 & 39.9 & 0.10 & 0.33 & [0.15, 0.52] \\
Basic AG-CoT & 99.9 & 99.9 & 2.62 & 5.37 & [4.67, 6.06] \\
Step AG-CoT & 99.9 & 99.9 & 2.31 & 4.83 & [4.15, 5.50] \\
GRPO-v2 & 99.9 & 99.9 & 2.21 & 5.26 & [4.57, 5.98] \\
RFT-v1 & 99.6 & 99.6 & \best{3.13} & \best{6.14} & [5.37, 6.91] \\
\bottomrule
\end{tabular}
}
\vspace{-0.5pt}
\end{wraptable}

\noindent\textbf{Surface validity saturates before exact target matching.}
Zero-shot outputs parse and remain Clifford-valid less than half the time, whereas AG-CoT SFT nearly saturates both checks.
The trained models thus learn the output language and Clifford gate constraints more readily than the target transformation, with most valid circuits still preparing the wrong state.
Appendix Table~\ref{tab:failure-breakdown} gives the complete strict-label failure categories.

\noindent\textbf{Verifier-filtered continuation gives the strongest direct result in the 32B study.}
RFT-v1 achieves the highest observed StateEq@1 among the reported 32B direct variants, at 6.14\%.
It also gives the highest strict saved-label Exact@1 among these variants, at 3.13\%.
Neither the intermediate row supervision in Step AG-CoT nor the reward tuning in GRPO-v2 exceeds Basic AG-CoT on either direct target-match metric.
GRPO-v2 strict Exact@1 is 2.21\%, with successes concentrated among shallow reference circuits, as detailed in Appendix Table~\ref{tab:depth-breakdown}.
It solves 35 of 47 targets with one or two reference-circuit layers, but only 1 of 2,743 targets with more than 50 layers.
Its mean training reward rises from 0.088 early in epoch 1 to 0.197 late in epoch 3, while its direct saved-label rate remains below Basic AG-CoT's 2.62\%.
The partial-label training reward and exact held-out match therefore measure different outcomes.

\begin{wraptable}{r}{.585\linewidth}
    \caption{Pass@$N$ (\%) from 64 candidates per target. Values for $N<64$ are estimated from this pool, and the $N=64$ column gives observed coverage. The Basic AG-CoT rows use the same 249,216 candidates, with StateEq-Pass@64 of 10.07\% versus 4.42\% under strict saved label equality.}
    \label{tab:appendix-passn-coverage}
    \resizebox{.98\linewidth}{!}{ 
    \begin{tabular}%
    {@{\hspace{.1mm}}l@{\hspace{.951mm}} @{\hspace{.51mm}}l@{\hspace{.591mm}} @{\hspace{.51mm}}c@{\hspace{.951mm}} @{\hspace{.51mm}}c@{\hspace{.951mm}}@{\hspace{.51mm}}c@{\hspace{.951mm}} @{\hspace{.51mm}}c@{\hspace{.591mm}} @{\hspace{.51mm}}c@{\hspace{.951mm}} @{\hspace{.951mm}}c@{\hspace{.1mm}}}
    \toprule
    Model & Criterion & 1 & 4 & 8 & 16 & 32 & 64 \\
    \midrule
    RFT-v1 & Label & 3.18 & 3.69 & 4.00 & 4.37 & 4.78 & 5.19 \\
    Basic AG-CoT & Label & 2.13 & 2.93 & 3.44 & 3.78 & 4.01 & 4.42 \\
    Basic AG-CoT & StateEq & 4.65 & 6.40 & 7.33 & 8.29 & 9.24 & 10.07 \\
    \bottomrule
    \end{tabular}
    }
\end{wraptable}
\noindent\textbf{State equivalence removes representation penalties, and verifier selection increases search coverage.}
State equivalence still requires the exact target state, but does not require the same generator basis.
Replacing saved-label equality with state equivalence roughly doubles the direct rate of every trained row in Table~\ref{tab:main-results}.
For Basic AG-CoT, we evaluate the same generated candidates under saved-label equality and state equivalence.
With 64 candidates per target, strict saved-label Pass@64 is 4.42\%, while StateEq-Pass@64 is 10.07\%, a 2.28$\times$ increase in measured coverage after removing generator-basis sensitivity.
For these candidates, estimated StateEq-Pass@1 is 4.65\%, while observed StateEq-Pass@64 is 10.07\%.
This is finite-budget search coverage rather than a direct-generation rate.
Table~\ref{tab:appendix-passn-coverage} gives the complete budget comparison for these pools, including the complementary RFT results.
Under stochastic decoding at temperature 0.8, RFT reaches observed saved-label Pass@64 of 5.19\%, compared with estimated Pass@1 of 3.18\% from the same pool.
The two Basic AG-CoT rows share candidates and differ only in the terminal criterion, whereas the RFT row uses a different checkpoint.
They therefore separate the effect of candidate budget from the choice of success criterion, rather than form a single ranking of training methods.

\subsection{Error Analysis}
\label{sec:verifier-diagnostics}

Exact success counts only fully correct circuits.
To test whether trace supervision also changes the incorrect outputs, we measure how many independent target constraints each incorrect output misses.
For each target-output pair, we form the subgroup generated by signed stabilizers shared by both states.
The rank deficit is the target stabilizer rank minus the rank of this shared subgroup.
It counts independent target constraints absent from the shared subgroup and is invariant to the generator basis.
The deficit measures shared state constraints, not the number of gates needed for circuit repair.

\begin{wraptable}{r}{.61\linewidth}
\centering
\caption{Constraint deficits among valid but state-incorrect outputs of adapted Qwen2.5-3B models in Table~\ref{tab:3b-training-comparison}. The deficit counts missing independent target constraints. Each model has its own error set. Means, medians and percentages use the corresponding error count, not a common set of failed targets. Outputs with deficit one also contribute to the final column.}
\label{tab:3b-constraint-errors}
    \resizebox{.97\linewidth}{!}{ 
    \begin{tabular}%
    {@{\hspace{.1mm}}l@{\hspace{.951mm}} @{\hspace{.951mm}}c@{\hspace{.591mm}} @{\hspace{.51mm}}c@{\hspace{.951mm}} @{\hspace{.51mm}}c@{\hspace{.951mm}}@{\hspace{.51mm}}r@{\hspace{.951mm}} @{\hspace{.51mm}}r@{\hspace{.1mm}} }
    \toprule
    System & Errors & \shortstack{Mean\\deficit} & \shortstack{Median\\deficit} & \shortstack{Deficit $= 1$\\Count (\%)} & \shortstack{Deficit $\leq 2$\\Count (\%)} \\
    \midrule
    Direct SFT & 3,616 & 8.22 & 9 & 39 (1.08) & 148 (4.09) \\
    AG-CoT SFT & 3,425 & 7.66 & 8 & 103 (3.01) & 266 (7.77) \\
    RFT ver. 1 & 3,398 & 7.61 & 8 & 121 (3.56) & 278 (8.18) \\
    RFT ver. 2 & 3,409 & 7.60 & 8 & 108 (3.17) & 279 (8.18) \\
    RFT ver. 3 & 3,416 & 7.57 & 8 & 115 (3.37) & 274 (8.02) \\
    \bottomrule
    \end{tabular}
    }%
\end{wraptable}

\noindent\textbf{Low-deficit errors are more common with trace supervision.}
Table~\ref{tab:3b-constraint-errors} applies the same constraint-deficit measure to the Qwen single-output evaluations in Section~\ref{sec:3b-training-results}.
We include only state-incorrect outputs with valid syntax, Clifford validity, the requested gate set, and the correct qubit count.
Among the outputs, the median deficit is 9 for Direct SFT and 8 for all AG-CoT SFT versions.
The share lacking at most two independent target constraints is 4.09\% for Direct SFT and 7.77\% for AG-CoT SFT, with the RFT versions at 8.02--8.18\%.
The largest difference is between circuit-only and trace supervision, while the additional change after RFT is smaller.
Alongside the exact successes in Table~\ref{tab:3b-training-comparison}, these distributions show a larger share of low-deficit outputs among the remaining valid errors. %
Mean deficit decreases from 8.22 for Direct SFT to 7.66 for AG-CoT and 7.57$\sim$7.61 for RFT.
The fraction missing exactly one constraint rises from 1.08\% for Direct SFT to 3.01\% for AG-CoT and 3.17\%$\sim$3.56\% for RFT.
These comparisons use each model's own valid errors, not a common subset of failed targets.

Appendix~\ref{app:32b-reference-comparisons} shows that the 32B RFT-v1 errors also retain target information.
Among 961 valid errors at five to eight qubits, 112 lack at most two independent target constraints, compared with 8.84 expected for random programs that preserve gate types and order.

\section{Conclusion}

This paper shows that algorithmic traces and exact verification together improve the semantic correctness of language-model Clifford circuit synthesis.
Because Clifford circuits admit exact classical verification, one verifier checks the traces used for supervision, selects model outputs for continued training, and scores generated circuits.
On 3,661 test targets, Qwen2.5-3B-Instruct with AG-CoT SFT yields 201 correct single outputs (5.49\%), compared with 33 (0.90\%) for circuit-only SFT.
On the same targets, Mistral-7B-Instruct-v0.3 improves from 71 correct outputs with Direct SFT to 316 with AG-CoT SFT.
Verifier-filtered supervised continuation gives a smaller observed increase, yielding 210--215 correct outputs across the three Qwen RFT versions (5.74--5.87\%) and 322 for Mistral.
Among the Qwen models' remaining valid errors, trace-trained models also have a higher proportion of outputs lacking at most two independent target constraints.
The complementary 32B study finds that RFT gives the highest direct StateEq@1 (6.14\%), while GRPO-v2 does not improve strict saved-label correctness.
Verifier-selected search finds additional correct circuits, with coverage reflecting multiple candidates per target rather than single-output correctness.
These results support algorithmic supervision and verifier-filtered continuation as complementary ways to improve semantic correctness, which code validity alone does not establish.

\subsection*{AI use statement}
A large language model assisted the authors in polishing the language of the manuscript. The authors verified that all claims, proofs, mathematical formulations, and reported values are valid, checked them against the implementation and results, and take full responsibility for the manuscript.

\bibliographystyle{iclr2027_conference}
\bibliography{references}

\appendix

\section{Formal Task and Verifier}
\label{sec:appendix-formal-task-verifier}

\begin{samepage}
This appendix formalizes the verifier-centered Clifford state-preparation task used throughout the paper.
The experiments instantiate the task using Clifford state-preparation targets from QCircuitBench.
For each target instance $x$, the prompt $p_x$ contains signed Pauli stabilizer generators that specify a target stabilizer-state tableau $T_x$.
The model emits a text program $y$ in OpenQASM 3.0.
The verifier treats $y$ as a candidate state-preparation circuit from the initial state $|0\rangle^{\otimes n}$.
The strict 32B criterion compares saved stabilizer labels, while StateEq and the controlled task-success rule compare the represented states.
The concrete implementation extracts a fenced \texttt{qasm} block when present, parses the candidate with Qiskit, converts the parsed circuit to a Qiskit \texttt{Clifford}, and constructs \texttt{StabilizerState(Clifford(circuit))} before applying the success criterion of each comparison.
\end{samepage}

\begin{table*}[htbp]
\centering
\small
\setlength{\tabcolsep}{5pt}
\caption{Verifier protocol for the Clifford state-preparation task.}
\label{tab:appendix-formal-task-boundary}
\begin{tabular*}{\textwidth}{@{\extracolsep{\fill}}p{0.18\textwidth}p{0.30\textwidth}p{0.40\textwidth}@{}}
\toprule
Component & Object & Verifier role \\
\midrule
Evaluation source & QCircuitBench Clifford random-circuit instances & Provides targets with exact stabilizer-tableau verification. \\
Prompt target & Signed Pauli generators $T_x$ & Specifies the target state without exposing a reference circuit. \\
Candidate output & OpenQASM 3.0 program $y$ & Defines an executable state-preparation circuit from $|0\rangle^{\otimes n}$. \\
Strict 32B criterion & $P(y)$, $K(y)$, and $\mathcal{L}(T_y)=\mathcal{L}(T_x)$ & Measures saved-label Exact@1. \\
State-equivalence criterion & $P(y)$, $K(y)$, and $\mathrm{Equiv}(T_y,T_x)=1$ & Removes generator-basis sensitivity. \\
Controlled task criterion & StateEq plus gate, qubit, header, and completion requirements & Measures task success in Table~\ref{tab:3b-training-comparison}. \\
Search decision & $\max_{1\leq i\leq N}\mathrm{Exact}(x,y_i)$ & Measures finite-budget verifier-selected coverage, not one-shot reliability. \\
\bottomrule
\end{tabular*}\par\vspace{-0.5pt}
\end{table*}

Let $n$ be the number of qubits, and let $X_j$, $Y_j$, and $Z_j$ denote the Pauli operators acting on qubit $j$.
Let $\mathrm{i}$ denote the imaginary unit.
For generator index $i$, let $r_i\in\{0,1\}$ be the sign bit, and let $a_{ij},b_{ij}\in\{0,1\}$ be the binary $X$ and $Z$ support bits on qubit $j$.
A stabilizer state $|\psi\rangle$ is represented by $n$ independent commuting signed Pauli generators $S_1,\ldots,S_n$ satisfying $S_i|\psi\rangle=|\psi\rangle$ for every generator index $i$.
Each generator is written as
\[
    S_i =
    (-1)^{r_i}
    \prod_{j=1}^{n}
    \left(\mathrm{i}^{a_{ij}b_{ij}} X_j^{a_{ij}} Z_j^{b_{ij}}\right),
\]
where the factor on qubit $j$ is $Y_j$ when both support bits are one, since $Y_j=\mathrm{i}X_jZ_j$.
The target tableau is $T_x=(M_x,r_x)$, where $M_x\in\{0,1\}^{n\times 2n}$ stores the binary $X/Z$ support and $r_x\in\{0,1\}^{n}$ stores the signs \citep{aaronson2008improvedsimulationstabilizercircuits}.
The implementation represents $T_x$ as Qiskit signed stabilizer labels. The mathematical tableau notation describes the same finite stabilizer-state object.

The verifier is an ordered decision procedure.
Let $P(y)$ be 1 when $y$ parses as OpenQASM and 0 otherwise.
Let $K(y)$ be 1 when the parsed circuit can be converted to Qiskit's Clifford representation and 0 otherwise. If $P(y)=0$, then $K(y)=0$.
When $P(y)=K(y)=1$, let the parsed circuit be $C_y=g_m\cdots g_2g_1$, where $g_k$ is the $k$-th gate in execution order.
Starting from the stabilizer tableau $T_{0^n}$ of $|0\rangle^{\otimes n}$, the prepared-state tableau is
\[
    T_y =
    F_{g_m}\!\left(\cdots F_{g_2}\!\left(F_{g_1}(T_{0^n})\right)\cdots\right),
\]
where $F_{g_k}$ is the exact tableau update induced by gate $g_k$.
The strict saved-label decision of the 32B study is
\[
    \mathrm{Exact}(x,y)=P(y)K(y)\mathbf{1}[\mathcal{L}(T_y)=\mathcal{L}(T_x)],
\]
where $\mathbf{1}[\cdot]$ equals 1 when the condition is true and 0 otherwise, and $\mathcal{L}(T)$ denotes the set of Qiskit signed stabilizer labels for tableau $T$.
StateEq replaces saved-label equality with $\mathrm{Equiv}(T_y,T_x)=1$ after the same parsing and Clifford-conversion stages.
The controlled task-success criterion additionally applies the gate, qubit, header, and completion requirements.

The verifier applies the stages in a fixed order: extraction, parsing, Clifford-family checking, tableau execution, and equality.
The extractor first prefers a fenced \texttt{qasm} block, then a generic fenced code block, and finally text beginning with \texttt{OPENQASM}.
A parse failure sets all semantic metrics to zero; a Clifford-conversion failure receives syntax credit only; the prepared tableau $T_y$ is computed only after both earlier stages pass.
The first failed stage determines the failure label.

For a generated program $y$, syntax success and Clifford validity are
\[
    \mathrm{Syntax}(y)=P(y),
    \qquad
    \mathrm{Valid}(y)=P(y)K(y).
\]
Exact@1 is the average of $\mathrm{Exact}(x,y)$ when one candidate is evaluated for each target.
For search-time evaluation, let $\mathcal{Y}_N(x)=\{y_1,\ldots,y_N\}$ be $N$ sampled candidates for the same target $x$.
The target-level Pass@$N$ indicator is
\[
    \mathrm{Pass@}N(x,\mathcal{Y}_N)
    =
    \max_{y_i\in\mathcal{Y}_N(x)}
    \mathrm{Exact}(x,y_i).
\]
This indicator defines success for a particular set of $N$ candidates.
The reported search curves estimate the probability of this event from 64 candidates per target, rather than generating a separate pool for each budget.
Let $\mathcal{X}$ be the test target set, let $B=64$ be the number of generated candidates per target, and let $s_x$ count the candidates satisfying $\mathrm{Exact}(x,y)=1$ for target $x$.
For $1\leq N\leq B$, the reported estimate is \citep{chen2021evaluatinglargelanguagemodels}
\[
    \widehat{\mathrm{Pass@}N}
    =\frac{1}{|\mathcal{X}|}\sum_{x\in\mathcal{X}}
    \left(1-\frac{\binom{B-s_x}{N}}{\binom{B}{N}}\right),
\]
where $\binom{B-s_x}{N}=0$ when $B-s_x<N$.
The ratio is the fraction of candidate subsets of size $N$ containing only failures.
The estimate at $N=1$ averages the fraction of correct candidates per target, while the estimate at $N=64$ equals the observed target coverage.
StateEq-Pass@$N$ uses the same estimator, with $s_x$ counting candidates that pass StateEq rather than saved label equality.
These sampling estimates are reported separately from direct Exact@1, which evaluates one output per target in the direct runs.

When the analysis conditions on earlier verifier stages, let $\mathcal{I}$ be the set of evaluated target-output pairs $(x,y)$ for which $P(y)=K(y)=1$.
The conditional exact rate is
\[
    \mathrm{Exact\mid Valid}
    =
    \frac{\sum_{(x,y)\in \mathcal{I}}\mathbf{1}[\mathcal{L}(T_y)=\mathcal{L}(T_x)]}{|\mathcal{I}|}.
\]
This measures how often parser-valid Clifford circuits match the saved target stabilizer labels after malformed outputs have already been removed.
For search reporting, the search lift at budget $N$ is
\[
    \mathrm{Lift@}N
    =
    \mathrm{Pass@}N-\mathrm{Exact@1}.
\]
The lift separates sparse search-time coverage from reliable direct generation.

\section{Dataset and Preprocessing Protocol}
\label{sec:appendix-dataset-preprocessing}

This appendix describes the dataset and split protocol for the 32B study, using Clifford random-circuit instances from QCircuitBench.
Inputs are signed Pauli stabilizer generators, outputs are OpenQASM 3.0 circuits, and all reported evaluation uses the verifier in Appendix~\ref{sec:appendix-formal-task-verifier}.
The 32B split groups duplicate QASM strings rather than duplicate target tableaus.
The controlled Qwen and Mistral evaluations instead use canonical target identities to isolate test targets from training and development.
Non-Clifford random-circuit synthesis and full gate-count or depth distributions are outside the 32B study.

The conversion from numeric target representation to signed-generator target representation has three steps.
First, the reference OpenQASM circuit is parsed into a Clifford object.
Second, the stabilizer generators are extracted as signed Pauli strings.
Third, these signed Pauli strings become the prompt-side target, while OpenQASM remains the completion-side output language.
For the Clifford instances studied here, this conversion preserves the target representation.
A 500-case conversion check tested agreement between the saved target labels and the stabilizer state prepared by the reference circuit.
This check motivated separating the random-circuit data into a Clifford stabilizer-tableau task and a universal-circuit task. The universal task retains state-vector-style targets and is not evaluated by the exact-tableau verifier used here.

For a general pure quantum state, let $b$ be an $n$-bit computational-basis string, let $|b\rangle$ be its basis vector, and let $\alpha_b$ be the complex amplitude assigned to $|b\rangle$.
A state-vector target has the form
\[
    |\psi\rangle
    =
    \sum_{b\in\{0,1\}^{n}}\alpha_b|b\rangle .
\]
This representation contains $2^n$ complex amplitudes before text formatting.
For $n=12$, this is 4096 complex amplitudes, and the printed target in the conversion example was roughly 16K characters.
By contrast, the stabilizer-state tableau stores $2n^2+n$ bits before text formatting.
For the same 12-qubit example, the printed tableau prompt was roughly 800 characters.
Thus the preprocessing changes the model-facing target from an exponential-size numeric string to a compact exact Clifford-state representation.

\begin{figure}[!htbp]
\vspace{0.75pt}
\centering
\resizebox{0.94\textwidth}{!}{%
\begin{tikzpicture}[
    font=\scriptsize,
    panel/.style={draw=black!20, rounded corners=6pt, fill=white, align=center, inner sep=5pt, minimum height=2.65cm},
    title/.style={font=\bfseries\scriptsize, align=center},
    chip/.style={draw=black!18, rounded corners=3pt, fill=white, align=center, inner sep=3pt, minimum height=0.46cm},
    flow/.style={-{Latex[length=1.9mm]}, line width=0.75pt, draw=black!60},
    badge/.style={draw=black!18, rounded corners=3pt, fill=white, align=center, inner sep=2.5pt}
]
\node[panel, fill=llmRed!65, minimum width=3.25cm] (vector) at (0,0) {};
\node[title] at ($(vector.north)+(0,-0.27)$) {Numeric state vector};
\node[chip, text width=2.58cm] at ($(vector.center)+(0,0.50)$)
{$|\psi\rangle=\sum_b \alpha_b |b\rangle$};
\node[chip, text width=2.58cm] at ($(vector.center)+(0,-0.12)$)
{$2^n$ complex amplitudes};
\node[chip, text width=2.58cm, fill=white] at ($(vector.center)+(0,-0.76)$)
{12 qubits: $\sim$16K printed chars};

\node[panel, fill=llmBlue!65, minimum width=3.25cm, right=1.38cm of vector] (tableau) {};
\node[title] at ($(tableau.north)+(0,-0.27)$) {Signed stabilizer target};
\node[chip, text width=2.58cm] at ($(tableau.center)+(0,0.50)$)
{$T_x=(M_x,r_x)$};
\node[chip, text width=2.58cm] at ($(tableau.center)+(0,-0.12)$)
{$n$ signed Pauli generators};
\node[chip, text width=2.58cm, fill=white] at ($(tableau.center)+(0,-0.76)$)
{12 qubits: $\sim$800 prompt chars};

\draw[flow] ($(vector.east)+(0,0.18)$) -- node[badge, fill=llmOrange, text width=1.18cm, above=1pt]
{$\approx 20{\times}$\\shorter} ($(tableau.west)+(0,0.18)$);

\node[panel, fill=llmGreen!65, minimum width=4.20cm, right=0.55cm of tableau] (verifier) {};
\node[title] at ($(verifier.north)+(0,-0.25)$) {Exact Clifford-state verifier};
\node[badge, text width=3.45cm, fill=white] at ($(verifier.center)+(0,0.50)$) {LLM-readable prompt};
\node[badge, text width=3.45cm, fill=white] at ($(verifier.center)+(0,-0.12)$) {same target state};
\node[badge, text width=3.45cm, fill=white] at ($(verifier.center)+(0,-0.76)$) {deterministic check};
\draw[flow] (tableau.east) -- (verifier.west);
\end{tikzpicture}
}
\caption{Target representation shift. Signed stabilizer generators replace long numeric state vectors with a compact target interface that still supports exact Clifford-state verification. Here $n$ is the qubit count, $M_x$ stores binary $X/Z$ support, and $r_x$ stores generator signs.}
\label{fig:target-representation}
\par\vspace{1pt}
\end{figure}
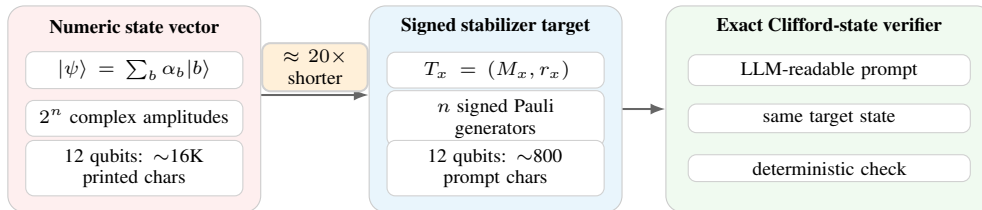

\begin{table}[H]
\centering
\small
\setlength{\tabcolsep}{5pt}
\caption{Representations used in the preprocessed Clifford state-preparation task.}
\label{tab:appendix-representation-boundary}
\begin{tabular*}{\textwidth}{@{\extracolsep{\fill}}p{0.27\textwidth}p{0.40\textwidth}p{0.25\textwidth}@{}}
\toprule
Representation & Model-facing object & Role in this study \\
\midrule
State vector & $2^n$ complex amplitudes & General but long text. \\
Stabilizer-state tableau & Signed Pauli generators; $2n^2+n$ bits before formatting & Compact exact Clifford target. \\
OpenQASM output & Clifford circuit program & Executable candidate. \\
\bottomrule
\end{tabular*}\par\vspace{-9pt}
\end{table}

The split is deterministic and controls duplicate completions at the \texttt{qasm\_output} string level.
The preprocessing uses a 90/10 train-test split with seed 42, groups identical \texttt{qasm\_output} strings before splitting, and uses no validation split.
The full accepted AG-CoT dataset contains 38,940 examples with no generation failures.
After splitting, the AG-CoT training set contains 35,046 examples, and the held-out Clifford test set contains 3,894 prompts.

\begin{table}[H]
\vspace{-0.75pt}
\centering
\footnotesize
\setlength{\tabcolsep}{4pt}
\caption{Dataset construction and validation counts.}
\label{tab:appendix-dataset-counts}
\begin{tabular*}{\columnwidth}{@{\extracolsep{\fill}}p{0.40\columnwidth}rp{0.38\columnwidth}@{}}
\toprule
Split or check & Count & Notes \\
\midrule
Accepted AG-CoT examples & 38,940 & Before train/test split. \\
Generation failures & 0 & During AG-CoT construction. \\
Training examples & 35,046 & 90\% split. \\
Held-out test prompts & 3,894 & 10\% split. \\
Conversion check & 500 & Random tableau/circuit checks. \\
AG validation & 1,000 & Random Clifford tests. \\
\bottomrule
\end{tabular*}\par\vspace{-5pt}
\end{table}

\begin{table}[H]
\centering
\small
\setlength{\tabcolsep}{4pt}
\caption{Held-out test prompt counts by qubit count. The counts sum to 3,894.}
\label{tab:appendix-test-qubit-counts}
\begin{tabular}{rrrrrrrrrrrr}
\toprule
Qubits & 2 & 3 & 4 & 5 & 6 & 7 & 8 & 9 & 10 & 11 & 12 \\
\midrule
Prompts & 30 & 51 & 107 & 145 & 219 & 311 & 367 & 475 & 618 & 702 & 869 \\
\bottomrule
\end{tabular}
\end{table}

\section{AG-CoT Algorithm Details}
\label{sec:ag-cot-appendix}

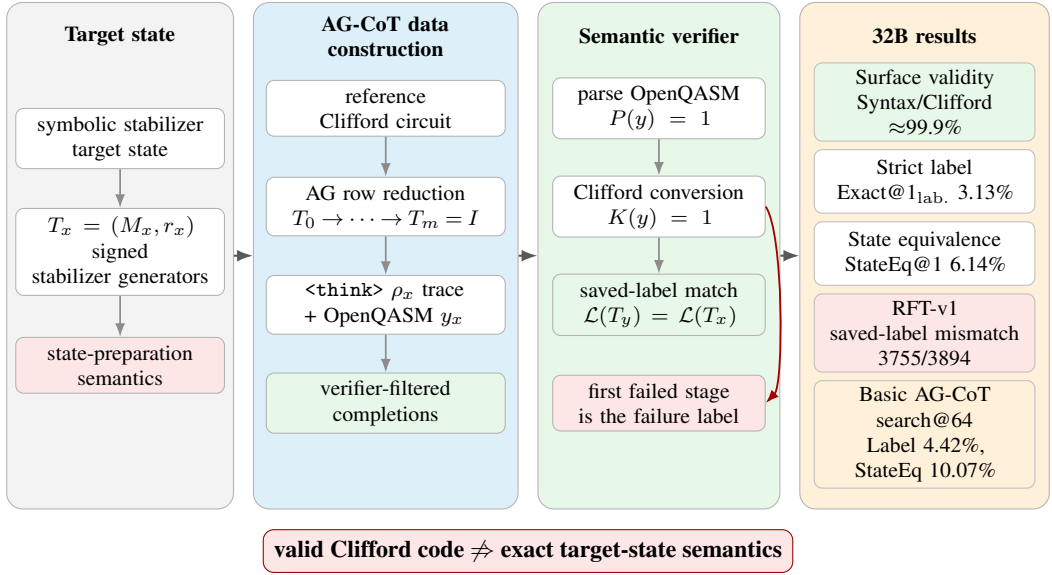
\begin{figure*}[!htbp]
\vspace{1pt}
\centering
\setlength{\abovecaptionskip}{1.75pt}
\begin{tikzpicture}[
    font=\fontsize{8}{9.5}\selectfont,
    card/.style={draw=black!30, rounded corners=6pt, minimum height=6.70cm, inner sep=5pt, align=center},
    title/.style={font=\bfseries\fontsize{8}{9.5}\selectfont, align=center, text width=2.85cm},
    chip/.style={draw=black!30, rounded corners=3pt, fill=white, align=center, inner sep=3pt, minimum height=0.58cm},
    metric/.style={draw=black!25, rounded corners=3pt, fill=white, align=center, inner sep=2.4pt, text width=2.75cm, minimum height=0.84cm},
    flow/.style={-{Latex[length=2.1mm]}, line width=0.8pt, draw=black!65},
    softflow/.style={-{Latex[length=1.8mm]}, line width=0.65pt, draw=black!45},
    failflow/.style={-{Latex[length=1.8mm]}, line width=0.7pt, draw=red!65!black}
]
\node[card, fill=llmGray, minimum width=3.00cm] (target) at (0,0) {};
\node[card, fill=llmBlue, minimum width=3.50cm, right=0.25cm of target] (cot) {};
\node[card, fill=llmGreen, minimum width=3.20cm, right=0.25cm of cot] (verify) {};
\node[card, fill=llmOrange, minimum width=3.30cm, right=0.25cm of verify] (result) {};

\node[title] at ($(target.north)+(0,-0.45)$) {Target state};
\node[chip, text width=2.55cm] (t1) at ($(target.center)+(0,1.55)$) {symbolic stabilizer\\target state};
\node[chip, text width=2.55cm, fill=white] (t2) at ($(target.center)+(0,0.05)$) {$T_x=(M_x,r_x)$\\signed\\stabilizer generators};
\node[chip, text width=2.55cm, fill=llmRed] (t3) at ($(target.center)+(0,-1.45)$) {state-preparation\\semantics};
\draw[softflow] (t1) -- (t2);
\draw[softflow] (t2) -- (t3);

\node[title] at ($(cot.north)+(0,-0.45)$) {AG-CoT data construction};
\node[chip, text width=2.95cm] (c1) at ($(cot.center)+(0,1.95)$) {reference\\Clifford circuit};
\node[chip, text width=2.95cm] (c2) at ($(cot.center)+(0,0.65)$) {AG row reduction\\$T_0 \!\to\! \cdots \!\to\! T_m\!=\!I$};
\node[chip, text width=2.95cm, fill=white] (c3) at ($(cot.center)+(0,-0.65)$) {\texttt{<think>} $\rho_x$ trace\\+ OpenQASM $y_x$};
\node[chip, text width=2.95cm, fill=llmGreen] (c4) at ($(cot.center)+(0,-1.95)$) {verifier-filtered completions};
\draw[softflow] (c1) -- (c2);
\draw[softflow] (c2) -- (c3);
\draw[softflow] (c3) -- (c4);

\node[title] at ($(verify.north)+(0,-0.45)$) {Semantic verifier};
\node[chip, text width=2.60cm] (v1) at ($(verify.center)+(0,1.95)$) {parse OpenQASM\\$P(y)=1$};
\node[chip, text width=2.60cm] (v2) at ($(verify.center)+(0,0.65)$) {Clifford conversion\\$K(y)=1$};
\node[chip, text width=2.60cm, fill=llmGreen] (v3) at ($(verify.center)+(0,-0.65)$) {saved-label match\\$\mathcal{L}(T_y)=\mathcal{L}(T_x)$};
\node[chip, text width=2.60cm, fill=llmRed] (v4) at ($(verify.center)+(0,-1.95)$) {first failed stage\\is the failure label};
\draw[softflow] (v1) -- (v2);
\draw[softflow] (v2) -- (v3);
\draw[failflow] (v2.east) .. controls +(0.24,-0.35) and +(0.24,0.25) .. (v4.east);

\node[title] at ($(result.north)+(0,-0.45)$) {32B results};
\node[metric, fill=llmGreen, anchor=north] (r1) at ($(result.north)+(0,-0.80)$) {Surface validity\\Syntax/Clifford $\approx$99.9\%};
\node[metric, fill=white, below=0.10cm of r1] (r2) {Strict label\\Exact@1$_{\mathrm{lab.}}$ 3.13\%};
\node[metric, fill=white, below=0.10cm of r2] (r3) {State equivalence\\StateEq@1 6.14\%};
\node[metric, fill=llmRed, below=0.10cm of r3] (r4) {RFT-v1\\saved-label mismatch\\3755/3894};
\node[metric, fill=llmOrange, below=0.10cm of r4] (r5) {Basic AG-CoT search@64\\Label 4.42\%,\\StateEq 10.07\%};

\draw[flow] (target.east) -- (cot.west);
\draw[flow] (cot.east) -- (verify.west);
\draw[flow] (verify.east) -- (result.west);
\node[draw=red!60!black, fill=llmRed, rounded corners=4pt, font=\bfseries\footnotesize, inner sep=4pt] (takeaway) at ($(target.south west)!0.5!(result.south east)+(0,-0.55)$)
{valid Clifford code $\not\Rightarrow$ exact target-state semantics};
\end{tikzpicture}
\caption{Overview of the 32B construction and verification. The target $T_x=(M_x,r_x)$ stores binary Pauli support $M_x$ and generator signs $r_x$. The trace $\rho_x$ records $m$ reduction steps through full reference Clifford tableaus $T_0,\ldots,T_m$ to the identity $I$, and $y_x$ is the emitted OpenQASM program. For a generated program $y$, $P(y)$ and $K(y)$ indicate parsing and Clifford validity, $T_y$ is its prepared state tableau, and $\mathcal{L}$ extracts saved stabilizer labels. The rightmost panel reports 32B direct results and basic AG-CoT search with 64 candidates per target.}
\label{fig:verifier-pipeline}
\par\vspace{1pt}
\end{figure*}

\noindent\textbf{Reference-circuit recipe.}
This appendix specifies the data-construction procedure used for AG-CoT.
For each target instance $x$, the construction starts from the QCircuitBench reference OpenQASM circuit to form the Clifford object and recover the target stabilizer tableau $T_x$.
It then constructs a new supervised completion by running Aaronson-Gottesman row reduction on the parsed Clifford representation.
For multi-qubit instances, the final OpenQASM block serializes the inverse of the accumulated reduction circuit rather than copying the reference circuit.
Figure~\ref{fig:verifier-pipeline} summarizes this construction and its evaluation results.

Let $p_x$ be the prompt, $T_x$ be the target tableau, and $C_x^{\mathrm{ref}}$ be the reference circuit.
The generator first parses $C_x^{\mathrm{ref}}$ into a Clifford object and extracts $T_x$.
Let $\mathrm{AG}$ denote the deterministic construction routine, $G_x$ its returned preparation circuit, and $\rho_x$ its row-reduction trace.
The routine returns
\[
    (G_x,\rho_x)=\mathrm{AG}(C_x^{\mathrm{ref}}),
\]
For multi-qubit instances, let $R_x$ denote the circuit containing the row-reduction operations in their application order, followed by any phase corrections.
These operations reduce the reference Clifford tableau to the identity.
The generator obtains the preparation circuit by reversing their order and inverting each gate:
\[
    G_x=R_x^{-1}.
\]
Basic AG-CoT records each qubit's signed destabilizer row before reducing that qubit, followed by its reduction operations.
Step AG-CoT also records the corresponding destabilizer row after each operation within that qubit's reduction.
Any final phase corrections appear as a separate list of operations in both formats.
Thus $\rho_x$ records the reduction, while $G_x$ runs in the opposite direction to prepare the target state.

For a single-qubit instance, the construction instead calls the direct Clifford synthesizer and records the line \texttt{q0: 1-qubit direct decomposition}.
This branch returns the preparation circuit directly, without a per-operation reduction trace.
In either branch, the emitted program $y_x$ is
\[
    y_x=\mathrm{QASM}(G_x),
\]
where $\mathrm{QASM}$ serializes the preparation circuit into OpenQASM.
The supervised completion $c_x$ is
\[
    c_x=\mathrm{Format}(\rho_x,G_x,y_x),
\]
where $\mathrm{Format}$ writes the row-reduction trace inside the \texttt{<think>} block and appends $y_x$ as the completion output.
\par

\begin{figure*}[!htbp]
\vspace{1pt}
\centering
\begin{tikzpicture}[
    font=\footnotesize,
    stage/.style={draw=black!30, rounded corners=5pt, fill=white, align=center, text width=0.170\textwidth, inner sep=3pt, minimum height=2.0cm},
    title/.style={font=\bfseries\footnotesize},
    smallchip/.style={draw=black!25, rounded corners=3pt, fill=white, align=center, inner sep=3pt},
    flow/.style={-{Latex[length=2mm]}, line width=0.75pt, draw=black!60}
]
\node[stage, fill=llmGray] (parse) at (0,0) {
    \textbf{Parse reference}\\[2pt]
    $C_x^{\mathrm{ref}}\rightarrow$ Clifford\\
    recover $T_x$
};
\node[stage, fill=llmBlue, right=0.20cm of parse] (reduce) {
    \textbf{AG row reduction}\\[2pt]
    $(G_x,\rho_x)=\mathrm{AG}(C_x^{\mathrm{ref}})$\\
    reference Clifford $\to$ identity
};
\node[stage, fill=llmGreen, right=0.20cm of reduce] (format) {
    \textbf{Format completion}\\[2pt]
    $y_x=\mathrm{QASM}(G_x)$\\
    $\begin{aligned}c_x&=\mathrm{Format}(\rho_x,\\&\quad G_x,y_x)\end{aligned}$
};
\node[stage, fill=llmOrange, right=0.20cm of format] (filter) {
    \textbf{Verifier filter}\\[2pt]
    parse, Clifford, exact match\\
    retain completions with $V(x,y_x)=1$
};
\node[stage, fill=llmRed, right=0.20cm of filter] (train) {
    \textbf{Training pair}\\[2pt]
    $(p_x,c_x)\in\mathcal{D}_{\mathrm{AG}}^{32\mathrm{B}}$\\
    trace + OpenQASM circuit
};
\draw[flow] (parse) -- (reduce);
\draw[flow] (reduce) -- (format);
\draw[flow] (format) -- (filter);
\draw[flow] (filter) -- (train);

\node[smallchip, fill=llmGray, below=0.45cm of reduce, text width=4.0cm] (basic)
{\textbf{Basic AG-CoT:} one trace line per qubit with current destabilizer row and applied gates.};
\node[smallchip, fill=llmGray, below=0.45cm of filter, text width=4.0cm] (step)
{\textbf{Step AG-CoT:} intermediate destabilizer-row states after each reduction operation.};
\draw[flow, draw=black!35] (reduce.south) -- (basic.north);
\draw[flow, draw=black!35] (reduce.south east) .. controls +(0.8,-0.6) and +(-0.8,0.6) .. (step.north west);
\end{tikzpicture}
\caption{AG-CoT data construction for the 32B study. The construction records the row-reduction trace and the synthesized OpenQASM circuit from the same process, then retains completions whose circuits pass the strict saved-label verifier.}
\label{fig:ag-cot-construction}
\par\vspace{-0.5pt}
\end{figure*}
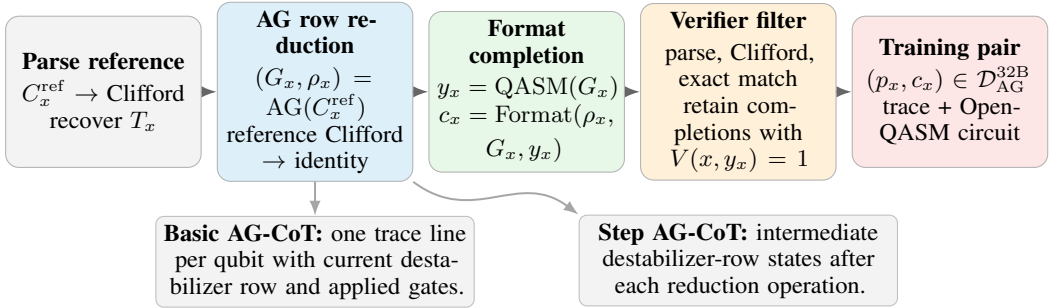

The verifier filter is the same semantic decision rule used for evaluation:
\[
    \begin{aligned}
    V(x,y_x)
    &=
    P(y_x)K(y_x)\\
    &\quad\cdot
    \mathbf{1}[\mathcal{L}(T_{y_x})=\mathcal{L}(T_x)],
    \end{aligned}
\]
where $V(x,y_x)=1$ means the emitted circuit passes parsing, Clifford-family checking, and exact target-tableau matching.
The accepted 32B AG-CoT dataset is
\[
    \mathcal{D}_{\mathrm{AG}}^{32\mathrm{B}}
    =
    \{(p_x,c_x):V(x,y_x)=1\}.
\]
This filtering retained 38,940 accepted examples with no observed generation failures.

The construction guarantee applies to the generated supervision.
The row-reduction implementation was checked against the expected Aaronson-Gottesman stages and validated on 1,000 random Clifford tests.
These tests validate the construction procedure, while verifier filtering accepts AG-CoT completions whose final OpenQASM circuits exactly prepare the target tableaus.
Generated model traces are evaluated through the emitted program.
The reported evaluation does not separately check every intermediate row in model-generated traces.
Thus the 32B AG-CoT training completions are trace-aligned and verifier-filtered, while learned model traces remain model outputs rather than validity certificates.

\noindent\textbf{Target-conditioned recipe.}
The Qwen2.5-3B-Instruct and matched Mistral comparisons use this construction.
Let $S_1,\ldots,S_n$ denote the signed stabilizer generators supplied in the prompt.
These $n$ rows specify the target state, but they do not specify a unique full Clifford operator.
The construction canonicalizes their generator basis and builds one deterministic full Clifford tableau $\widetilde{T}_x$ for the same state.
Its stabilizer half contains an equivalent basis for the target stabilizer group.
Its other $n$ rows are auxiliary destabilizers $D_1,\ldots,D_n$.
In the Clifford tableau convention, $D_i$ anticommutes with its paired stabilizer $S_i$ and commutes with $S_j$ for $j\neq i$ \citep{aaronson2008improvedsimulationstabilizercircuits}.
The destabilizer rows complete the symplectic frame required by the reduction.
They are not prompt fields and do not add target-state constraints.
The construction then reduces $\widetilde{T}_x$ to the identity, inverts the accumulated reduction, and compiles the resulting preparation circuit to \texttt{\{h, s, cx\}}.
The final circuit must parse, have the requested qubit count, use only the allowed gates, and prepare a state equivalent to the target.
The reference QASM is parsed separately to check source-target consistency and does not drive this teacher synthesis.

\section{Training and Search Interpretation}
\label{sec:appendix-training-search}

The training and selection methods in this paper use the verifier in different ways.
Each AG-CoT SFT model optimizes the conditional likelihood of its accepted completions:
\[
    \mathcal{L}_{\mathrm{SFT}}(\theta)
    =
    -\sum_{(p_x,c_x)\in \mathcal{D}_{\mathrm{AG}}^{(r)}}
    \log \pi_\theta(c_x\mid p_x),
\]
where $r$ identifies the reference-circuit or target-conditioned construction and $\pi_\theta$ is the fine-tuned language model with parameters $\theta$.

GRPO uses verifier-derived reward differently from SFT.
For a Clifford-valid output, let
\[
    m(x,y)=\frac{|\mathcal{L}(T_y)\cap\mathcal{L}(T_x)|}{|\mathcal{L}(T_x)|}
\]
denote the fraction of saved target labels that also appear in the generated label set.
The reward used in the reported GRPO-v2 run is
\[
R(x,y)=
\begin{cases}
1, & P(y)=1,\ K(y)=1,\ \mathcal{L}(T_y)=\mathcal{L}(T_x),\\
0.8\,m(x,y), & P(y)=1,\ K(y)=1,\ \mathcal{L}(T_y)\neq\mathcal{L}(T_x),\\
-0.5, & P(y)=1,\ K(y)=0,\\
-1, & P(y)=0.
\end{cases}
\]
Each update samples a group of $G=32$ outputs for one prompt and forms group-relative advantages from their rewards as in GRPO-style training \citep{shao2024deepseekmath}.
The mean training reward rises from 0.088 early in epoch 1 to 0.197 late in epoch 3.
Held-out strict saved-label Exact@1 is 2.21\%, compared with 2.62\% for Basic AG-CoT SFT, so the optimized reward and terminal held-out metric remain distinct quantities.

For the 32B study, rejection fine-tuning uses the strict saved-label verifier as a filter.
Let $\mathcal{S}(x)$ be the sampled set for target $x$.
Its RFT training set contains pairs
\[
    \mathcal{D}_{\mathrm{RFT}}^{32\mathrm{B}}
    =
    \{(x,y): y\in \mathcal{S}(x),\ \mathrm{Exact}(x,y)=1\}.
\]
\begin{samepage}
The 3B continuation instead uses the StateEq-based task-success rule specified in Appendix~\ref{sec:prompt-decoding-record}.
A strict saved-label mismatch does not reject a state-equivalent response that satisfies all 3B task requirements.
Best-of-$N$ uses the verifier for inference-time selection rather than parameter updates.
Thus Pass@64 = 5.19\% measures finite-budget search coverage: under 64 sampled candidates, verifier selection finds exact solutions for some targets, distinct from reliable direct generation.
\par
\end{samepage}

\section{Decoding and Evaluation Settings}
\label{sec:prompt-decoding-record}
\label{sec:decoding-evaluation-settings}

This appendix specifies the prompt format, decoding settings, sampling budgets, and verifier treatment used by the Clifford-synthesis evaluations.
The reported decoding configuration includes only parameters explicitly recorded for the evaluated runs.

\noindent\begin{minipage}{\linewidth}
\noindent\textbf{Prompt format.}
For the Clifford random-circuit task, each target is represented by stabilizer generators rather than a printed state vector.
The task prompt is:

\begin{PromptBlock}
Given an {n}-qubit state |\psi>
described by stabilizer generators:

{tableau}

Use gates {H, S, CNOT}.
Output OpenQASM for a circuit C such that:
C |0>^{\otimes {n}} = |\psi>.

Provide the circuit in OpenQASM 3.0 format.
\end{PromptBlock}
\end{minipage}

\noindent\begin{minipage}{\linewidth}
At inference time, the task prompt uses the following chat template:

\begin{PromptBlock}
Question: {prompt}
Answer:
\end{PromptBlock}
\end{minipage}

\noindent\begin{minipage}{\linewidth}
For supervised AG-CoT data, the assistant-side target contains a \texttt{<think>} trace followed by a fenced OpenQASM block:

\begin{PromptBlock}
<think>
... AG-CoT trace ...
</think>
```qasm
... OpenQASM 3.0 program ...
```
\end{PromptBlock}
\end{minipage}

\noindent\textbf{Output extraction and verifier treatment.}
Generated text is converted to a candidate circuit by first extracting a fenced \texttt{qasm} block, then a generic fenced code block, then text beginning with \texttt{OPENQASM}.
If none of these patterns is found, the raw text is passed to the verifier.
Invalid outputs are neither repaired nor discarded before scoring.
A parse failure receives zero for syntax, Clifford validity, and exact tableau match.
A parsed but non-Clifford circuit receives syntax credit only.
In the 32B study, Exact@1 requires equality between the expected saved stabilizer labels and the labels derived from the generated circuit.
StateEq@1 instead uses state equivalence after the same parsing and Clifford-conversion stages.
For the 3B comparison, task success requires StateEq together with the syntax, Clifford, gate-set, qubit-count, OpenQASM 3.0, and completion requirements stated in the 3B experimental setup.

\noindent\textbf{Evaluation run configuration.}
The 32B Best-of-64 evaluation covers 3,894 prompts with 64 samples at temperature 0.8 and uses the RFT-v1 checkpoint after its first training epoch.
The direct rows in that study use the same 3,894-prompt held-out set, as summarized in Table~\ref{tab:appendix-evaluation-budgets}.
The 32B Exact@1 values are computed from the evaluation runs in that table.
The 3B comparison uses 3,661 target-disjoint test prompts and one greedy completion per target, with temperature zero and top-$p$ set to 1.0.
Its maximum model length is 8,192 tokens and its nominal output cap is 8,000 tokens, reduced when necessary to fit the prompt in the context window.
These evaluations use no retries, output repair, or format conversion.

\noindent\textbf{Verifier environment.}
The 32B verifier environment used Qiskit 2.4.1, Qiskit Aer 0.15.1, and Stim 1.15.0.
The 3B evaluations use Qiskit 2.3.0 and Qiskit Aer 0.17.2.

\section{Reproducibility Details}
\label{app:artifact-reproducibility}

Reproducing the 32B study requires the 3,894 held-out evaluation inputs, verifier implementation, AG-CoT data-construction scripts, evaluation records, and runtime versions listed below.

\begin{table*}[htbp]
\centering
\small
\setlength{\tabcolsep}{5pt}
\caption{Components required to reproduce the 32B study.}
\label{tab:artifact-components}
\begin{tabular*}{\textwidth}{@{\extracolsep{\fill}}p{0.20\textwidth}p{0.36\textwidth}p{0.34\textwidth}@{}}
\toprule
Component & Required record & Reproducibility role \\
\midrule
Held-out prompts & 3,894 targets plus metadata & Replay held-out evaluation inputs. \\
Verifier/evaluator & Extraction, Clifford conversion, \mbox{Exact@1}, \mbox{StateEq@1}, Pass@$N$ & Recompute reported metrics. \\
AG-CoT construction & Row-reduction traces plus verifier acceptance & Rebuild SFT data. \\
Evaluation records & Commands, evaluated checkpoints, decoding settings, checksums & Identify the evaluated configuration. \\
Dependencies & Qiskit and model-stack versions & Recreate the software environment. \\
\bottomrule
\end{tabular*}\par\vspace{-0.75pt}
\end{table*}

\begin{samepage}
The evaluator first extracts an OpenQASM program from the model output, then parses it with Qiskit, converts the parsed circuit to a Clifford object, constructs \texttt{StabilizerState(Clifford(circuit))}, and compares the resulting saved stabilizer labels with the target labels.
StateEq@1 is computed using \texttt{StabilizerState.equiv} after the same parsing and Clifford-conversion stages.
Pass@$N$ uses the same exact-success indicator per sampled candidate and changes only the sample budget.
\par
\end{samepage}

The evaluation records specify the evaluated model, dataset split, decoding settings, verifier configuration, result schema, and software environment needed to reproduce the reported metrics.
The most important generation fields are sample count, maximum generated tokens, maximum model length, temperature, top-$p$, and stop-token policy.
A data manifest records the source dataset checksum, preprocessing script, split policy, task filters, record counts, and output checksums.

\begin{table}[htbp]
\centering
\footnotesize
\setlength{\tabcolsep}{3pt}
\caption{Software versions for the 32B study and the 3B single-output evaluations. A dash marks a package version not specified for the 3B evaluations.}
\label{tab:appendix-runtime-versions}
\begin{tabular*}{\columnwidth}{@{\extracolsep{\fill}}lcc@{}}
\toprule
Package & 32B study & 3B eval. \\
\midrule
Python & 3.11 & 3.12.3 \\
PyTorch & 2.9.1 & 2.5.1 \\
Transformers & 4.57.6 & 4.52.4 \\
TRL & 0.27.2 & -- \\
PEFT & 0.19.1 & -- \\
vLLM & 0.15.0 & 0.6.4.post1 \\
Accelerate & 1.13.0 & -- \\
BitsAndBytes & 0.49.2 & -- \\
Datasets & 4.8.5 & -- \\
Qiskit & 2.4.1 & 2.3.0 \\
Qiskit Aer & 0.15.1 & 0.17.2 \\
Stim & 1.15.0 & -- \\
\bottomrule
\end{tabular*}\par\vspace{-2pt}
\end{table}
\FloatBarrier

\subsection{Replication Across Inference Environments}

We evaluate the same AG-CoT SFT and RFT version 1 checkpoints on NVIDIA RTX A6000 and H200 NVL environments using the same 3,661 test prompts and evaluator.
Both evaluations use greedy FP16 inference, batches of 32, an 8,192 token context and a nominal 8,000 token output cap, with no output repair.
The A6000 environment uses CUDA 12.1 and Python 3.12.3, while the H200 environment uses CUDA 12.4 and Python 3.11.15.
The tokenizers and safetensors package versions also differ, so this comparison covers two software and hardware environments rather than an isolated hardware change.

Table~\ref{tab:3b-environment-replication} shows that exact task success is highly consistent across these evaluations.
All 201 AG-CoT successes on A6000 also succeed on H200, which solves one additional target.
RFT version 1 succeeds on exactly the same 213 targets in both environments.
Generated QASM is not identical across environments, so agreement in task success does not imply identical output text.

\begin{table}[htbp]
\centering
\small
\setlength{\tabcolsep}{5pt}
\caption{Exact task success across two inference environments on the same 3,661 targets. Shared successes count targets solved in both environments. The A6000 counts are those reported in Table~\ref{tab:3b-training-comparison}.}
\label{tab:3b-environment-replication}
\begin{tabular*}{\columnwidth}{@{\extracolsep{\fill}}lrrr@{}}
\toprule
System & A6000 & H200 NVL & Shared successes \\
\midrule
AG-CoT SFT & 201 & 202 & 201 \\
RFT version 1 & 213 & 213 & 213 \\
\bottomrule
\end{tabular*}\par\vspace{-11.5pt}
\end{table}
\FloatBarrier

\section{Additional Results and Compute}
\label{sec:appendix-additional-figures-tables}
\label{app:compute-runtime}

\subsection{Matched Mistral Training}
\label{app:mistral-training}

The Mistral comparison in Table~\ref{tab:3b-training-comparison} uses Mistral-7B-Instruct-v0.3 and the target-conditioned teacher corpus described in Appendix~\ref{sec:ag-cot-appendix}.
The two SFT datasets share prompts, target identities and final teacher circuits, with the algorithmic trace removed only from Direct SFT completions.
We exclude a training pair if either completion with its prompt exceeds 8,192 tokens under the Mistral tokenizer, retaining 35,005 of 35,046 pairs.
Both models train for one epoch with global batch size 32, learning rate $2\times10^{-4}$, BF16 precision and seed 42.
LoRA uses rank 32 and alpha 64, with dropout 0.05 on the query, key, value and output attention projections and the gate, up and down feedforward projections.
The comparison holds examples, updates and optimizer settings fixed, while trace supervision includes more completion tokens.

For continuation, the AG-CoT model generates one response for each of 8,192 training-pool prompts at temperature 0.8 and top-$p$ 0.95, with seed 42.
The verifier retains 983 complete responses whose emitted circuits satisfy the StateEq-based task requirements.
Their original text is retained without repair, including the generated trace.
Starting from the same AG-CoT adapter, RFT trains for one epoch with learning rate $5\times10^{-6}$, global batch size 32 and the same seed and LoRA settings.
Each training stage uses its final epoch checkpoint.

All three models are fixed before evaluation on the same 3,661 test targets used for the Qwen comparison.
Evaluation uses one greedy FP16 completion per target, batches of 16, an 8,192 token context and a nominal 8,000 token output cap reduced to fit each prompt.
There are no retries, output repair or test-based configuration changes.
Training uses PyTorch 2.5.1, Transformers 4.52.4 and PEFT 0.19.1, and evaluation uses vLLM 0.6.4.post1 and Qiskit 2.3.0.
All StateEq successes also satisfy the task requirements, giving the 71, 316 and 322 task successes in Table~\ref{tab:3b-training-comparison}.
In paired target comparisons, AG-CoT gains 246 successes and loses 1 relative to Direct SFT, with a two sided exact McNemar $p$ value of $2.19\times10^{-72}$.
RFT gains 24 successes and loses 18 relative to AG-CoT, with $p=0.441$, so its six additional successes constitute a small observed increase rather than a statistically established gain.
These are separate checkpoints and a different evaluation protocol from the 2,048-token development comparison in Table~\ref{tab:cross-family-direct-sft}.

\subsection{Paired Target Level Analysis}

Table~\ref{tab:paired-task-success} compares the Qwen2.5-3B-Instruct models' binary task success outcomes for the same targets used in Table~\ref{tab:3b-training-comparison}.
The Direct SFT to AG-CoT comparison has 172 discordant targets.
AG-CoT succeeds on 170 targets missed by Direct SFT, while Direct SFT succeeds on 2 targets missed by AG-CoT.
The RFT comparisons have smaller net gains.
After Holm adjustment across the three comparisons, version 2 remains below 0.05.

\subsection{Task Success by Qubit Count}

\begin{samepage}
Table~\ref{tab:3b-success-by-qubits} separates the Qwen Direct SFT and AG-CoT results by target size under the same single output protocol as Table~\ref{tab:3b-training-comparison}.
Each row includes all test targets at that size, including targets with invalid or incomplete outputs.
AG-CoT has more successes at every qubit count, so the aggregate gain is not confined to the smallest targets.
At twelve qubits, Direct SFT solves 2 of 841 targets and AG-CoT solves 40.
\par
\end{samepage}

\begin{table}[htbp]
\centering
\small
\setlength{\tabcolsep}{5pt}
\caption{Task success counts by qubit count for the 3B comparison. Both models use the same targets and one greedy completion per target. The target count is the denominator for both success rates in each row.}
\label{tab:3b-success-by-qubits}
\begin{tabular*}{\columnwidth}{@{\extracolsep{\fill}}rrrr@{}}
\toprule
Qubits & Targets & Direct SFT & AG-CoT SFT \\
\midrule
3 & 30 & 0 & 3 \\
4 & 95 & 1 & 8 \\
5 & 135 & 3 & 12 \\
6 & 201 & 2 & 10 \\
7 & 282 & 4 & 16 \\
8 & 355 & 7 & 29 \\
9 & 454 & 6 & 24 \\
10 & 596 & 4 & 27 \\
11 & 672 & 4 & 32 \\
12 & 841 & 2 & 40 \\
\midrule
All & 3,661 & 33 & 201 \\
\bottomrule
\end{tabular*}\par\vspace{-0.5pt}
\end{table}

\subsection{Constraint Deficits in Valid Errors}

Table~\ref{tab:3b-constraint-errors} summarizes the constraint deficits among the remaining valid errors of the 3B models.
The mean deficit is lower for AG-CoT and the RFT versions than for Direct SFT, consistent with the median and low deficit proportions discussed in Section~\ref{sec:verifier-diagnostics}.

\subsection{Matched Cross-Family Direct-SFT Replication}

Table~\ref{tab:cross-family-direct-sft} reports the matched Qwen and Mistral comparison.
Both Direct-SFT models use the same 35,046 target-to-QASM training instances.
The evaluation uses one greedy FP16 completion per row, a 2,048-token generation cap, and the same strict no-repair verifier.
The models retain their own tokenizers and model-specific LoRA target modules.
Direct SFT raises TaskSuccess@1 from 0 to 41 (1.12\%) for Qwen2.5-3B-Instruct and from 0 to 78 (2.13\%) for Mistral-7B-Instruct-v0.3.
The models differ in parameter count, tokenizer, and training implementation, so this experiment tests whether the direction of the Direct-SFT effect appears in another model family rather than ranking the two models.

\begin{table}[htbp]
\centering
\small
\setlength{\tabcolsep}{5pt}
\caption{Matched cross-family Direct-SFT replication on a common 3,661-row target-disjoint development partition. Each row uses one greedy FP16 completion with a 2,048-token generation cap and strict no-repair evaluation. These results are separate from the 8,000-token comparisons in Table~\ref{tab:3b-training-comparison}.}
\label{tab:cross-family-direct-sft}
\begin{tabular*}{\columnwidth}{@{\extracolsep{\fill}}llrr@{}}
\toprule
Family & Checkpoint & Correct / total & Task success (\%) \\
\midrule
Qwen & Base & 0 / 3,661 & 0.00 \\
Qwen & Direct SFT & 41 / 3,661 & 1.12 \\
Mistral & Base & 0 / 3,661 & 0.00 \\
Mistral & Direct SFT & 78 / 3,661 & 2.13 \\
\bottomrule
\end{tabular*}\par\vspace{-1.75pt}
\end{table}

The remainder of this appendix presents additional search results and evaluation costs from the 32B study.
Direct Exact@1 measures one output per target under that study's saved-label verifier convention.
The 32B search runs generate 64 candidates per target at temperature 0.8 and estimate Pass@$N$ for smaller budgets from the same candidates using the estimator in Appendix~\ref{sec:appendix-formal-task-verifier}.
At $N=64$, the estimate equals observed target coverage.
StateEq-Pass@$N$ uses the same estimator with state equivalence as the success criterion for each candidate.

Verifier-selected coverage rises with more samples, but the magnitude depends on both the checkpoint and the success criterion, as shown in Table~\ref{tab:appendix-passn-coverage}.
The strict row uses the verifier-filtered RFT checkpoint and saved-label equality.
The two Basic AG-CoT rows use the same generated candidates and differ only in the final success criterion, saved label equality versus state equivalence.
These rows compare search estimates under different criteria rather than providing a direct method ranking.
Under stochastic decoding at temperature 0.8, RFT reaches observed Pass@64 of 5.19\%, compared with estimated Pass@1 of 3.18\% from the same candidate pool.
For Basic AG-CoT, state equivalence increases measured 64-candidate coverage by 2.28 times relative to saved-label equality, although 89.93\% of prompts remain uncovered.

The same 64-sample budget is not equally useful across target sizes.
Coverage is highest for two-qubit prompts, but that bucket contains only 30 prompts. We use this stratification as a diagnostic of budget sensitivity rather than to fit a scaling law.

\begin{table}[htbp]
\centering
\small
\setlength{\tabcolsep}{2pt}
\caption{Search results by qubit count for RFT. Pass@1 averages the fraction of correct candidates among 64 samples per target, and Pass@64 gives observed target coverage.}
\label{tab:appendix-pass64-by-qubits}
\begin{tabular*}{\columnwidth}{@{\extracolsep{\fill}}rrrr@{}}
\toprule
Qubits & Prompts & Pass@1 (\%) & Pass@64 (\%) \\
\midrule
2  & 30  & 21.15 & 86.67 \\
3  & 51  & 7.20  & 21.57 \\
4  & 107 & 6.57  & 13.08 \\
5  & 145 & 4.70  & 7.59 \\
6  & 219 & 3.30  & 4.57 \\
7  & 311 & 7.11  & 8.04 \\
8  & 367 & 2.38  & 4.09 \\
9  & 475 & 2.46  & 3.79 \\
10 & 618 & 2.07  & 3.56 \\
11 & 702 & 2.87  & 3.99 \\
12 & 869 & 1.97  & 2.53 \\
\bottomrule
\end{tabular*}\par\vspace{-0.75pt}
\end{table}

Table~\ref{tab:appendix-evaluation-budgets} summarizes the 32B evaluation budgets on the same held-out set.
The Best-of-64 row verifies 249,216 generated candidates, and that run takes 9~h 52~min 49~s of wall-clock time.
The Basic AG-CoT search evaluation also covers 249,216 generated candidates over the same 3,894 held-out prompts.
These candidate counts specify the inference budget for each search evaluation. The wall-clock measurement describes the Best-of-64 evaluation, not average generation latency, verifier latency, or training time.

\begin{table}[htbp]
\centering
\footnotesize
\setlength{\tabcolsep}{2pt}
\caption{Evaluation budgets for the 32B results over 3,894 held-out prompts.}
\label{tab:appendix-evaluation-budgets}
\begin{tabular*}{\columnwidth}{@{\extracolsep{\fill}}p{0.25\columnwidth}cp{0.29\columnwidth}r@{}}
\toprule
Run & Samples & Metric & Value (\%) \\
\midrule
Zero-shot & 1 & Exact / StateEq & 0.10 / 0.33 \\
Basic AG-CoT & 1 & Exact / StateEq & 2.62 / 5.37 \\
Step \mbox{AG-CoT} & 1 & Exact / StateEq & 2.31 / 4.83 \\
GRPO-v2 & 1 & Exact / StateEq & 2.21 / 5.26 \\
RFT-v1 & 1 & Exact / StateEq & 3.13 / 6.14 \\
RFT-v1 Best-of-64 & 64 & Pass@64 & 5.19 \\
Basic AG-CoT Best-of-64 & 64 & \mbox{Pass@64 /} \mbox{StateEq-P@64} & 4.42 / 10.07 \\
\bottomrule
\end{tabular*}\par\vspace{-1.75pt}
\end{table}
\FloatBarrier

We also evaluated one prompt and output-format variant as an exploratory stress test.
This run uses the same 3,894 prompts with stabilizer-generator targets, while the supervised completion is direct OpenQASM rather than an AG-CoT trace followed by OpenQASM.
It also uses a shorter generation budget than the main rows.
For this QASM-only SFT model, the maximum generation length is 2,048 and the maximum model length is 16,384.
The prompt-token limit is 14,336, no prompt is skipped, and the maximum observed prompt length is 250 tokens.

\begin{table}[htbp]
\centering
\footnotesize
\setlength{\tabcolsep}{1pt}
\caption{Exploratory QASM-only format stress test under a 2,048-token generation cap.}
\label{tab:appendix-qasm-only-stress}
\begin{tabular*}{\columnwidth}{@{\extracolsep{\fill}}p{0.23\columnwidth}rrrrr@{}}
\toprule
Run & Max tok. & \multicolumn{4}{c}{Rate (\%)} \\
\cmidrule(l){3-6}
& & Syntax & Exact@1 & StateEq@1 & Trunc. \\
\midrule
QASM-only SFT & 2048 & 24.91 & 5.50 & 7.68 & 83.00 \\
\bottomrule
\end{tabular*}\par\vspace{-0.5pt}
\end{table}

The output-extraction and verifier rules are unchanged.
The 83\% truncation rate shows that the 2,048-token cap prevents complete outputs for most targets, so this run is a format-and-budget diagnostic rather than a directly comparable method ranking.

\FloatBarrier
\section{Target Constraint Retention in 32B Errors}
\label{app:32b-reference-comparisons}

We compare the 32B RFT-v1 errors with random references to determine whether they retain target information beyond that expected from program structure alone.
The analysis uses the 2,837 outputs whose saved QASM is shorter than the 2,000-character limit.
Of these, 2,581 form valid candidate states with the correct qubit count but fail state equivalence.
Together, the 2,837 outputs comprise 239 exact outputs, 1,246 valid outputs with nonzero fidelity but incorrect states, 1,335 valid outputs with zero fidelity and 17 syntax failures.
The constraint deficit defined in Section~\ref{sec:verifier-diagnostics} counts the independent signed target constraints absent from the shared stabilizer subgroup.

The program reference holds each recorded target fixed and preserves the corresponding generated program's ordered sequence of gate types.
It redraws the qubit operands uniformly and independently for each gate, with distinct operands for gates acting on two qubits.
We draw 60 programs per output using random seed 20260720 and exclude exact target matches, matching the exclusion of exact outputs from the analyzed errors.
For each output, the fraction of retained reference programs with deficit at most two estimates its reference probability.
Summing these probabilities gives the expected count in Table~\ref{tab:32b-reference-comparisons}.
The reference standard deviation is the square root of the sum of each probability multiplied by its complement.

Two additional references separate target structure from generated program structure.
The empty circuit emits no gates and prepares $|0\rangle^{\otimes n}$ on each recorded target's $n$ qubits.
Its deficit is evaluated directly on every target.
The uniform reference instead replaces the candidate with a uniformly random $n$-qubit stabilizer state, conditioned on not preparing the target exactly.
Its expectations are evaluated from the exact stabilizer-state distribution rather than from sampled circuits.
All three references keep the same targets and qubit counts, while only the length-matched reference preserves the generated program's gate sequence.

We report this comparison on the 961 errors at five to eight qubits.
No outputs at these sizes reach the 2,000-character limit.
This range also excludes two and three qubit targets, where small state spaces make low deficits less informative.
The observed count is 112, compared with an expected 8.84 and reference standard deviation 2.45.
Thus the generated errors contain 12.7 times as many outputs with deficit at most two as this reference predicts.
On the same 961 errors, the empty circuit gives 39 such outputs and the uniform reference expects 0.396.
Table~\ref{tab:32b-reference-strata} provides the complete comparison by qubit count, including the smaller and larger systems.
The 2,000-character limit removes no outputs at two to eight qubits, but removes 11 of 475, 208 of 618, 374 of 702 and 464 of 869 outputs at nine, ten, eleven and twelve qubits, respectively.
Because this exclusion acts on the generated candidate rather than the reference candidates, the larger-system rows describe the retained outputs and do not extend the five-to-eight-qubit comparison to all outputs.

A separate reference tests sign agreement over all 2,581 valid errors.
Let $m$ denote the dimension of the shared stabilizer subspace after removing Pauli signs.
Holding this unsigned overlap fixed for each output, the reference assigns independent fair relative signs to a basis of the shared subspace.
The probability of at least one sign conflict is then $1-2^{-m}$.
Summing these probabilities gives 1,639.2 expected conflicts with standard deviation 16.5, compared with 1,335 observed.
The generated signs therefore agree with the target more often than this reference predicts at the overlaps they achieve.
This sign reference conditions on the observed unsigned overlap and is distinct from the reference that randomizes gate operands.

\begin{table}[htbp]
\centering
\small
\setlength{\tabcolsep}{4pt}
\caption{Random reference comparisons for 32B RFT-v1 errors. The program reference preserves gate types and order on five to eight qubit targets. The sign reference preserves each output's unsigned overlap on all valid errors. The final column is the reference standard deviation, not a confidence interval.}
\label{tab:32b-reference-comparisons}
\begin{tabular*}{\columnwidth}{@{\extracolsep{\fill}}lrrrr@{}}
\toprule
Event & Errors & Observed & Expected & Std. dev. \\
\midrule
Deficit at most two & 961 & 112 & 8.84 & 2.45 \\
Sign conflict & 2,581 & 1,335 & 1,639.2 & 16.5 \\
\bottomrule
\end{tabular*}
\end{table}

\FloatBarrier
\begin{table}[!t]
\centering
\setlength{\tabcolsep}{4pt}
\caption{Complete reference comparison for the 2,581 valid, state-incorrect 32B RFT-v1 outputs. Here $n$ is the qubit count, $N$ is the number of analyzed errors, and $r$ counts missing independent signed target constraints. Empty denotes the zero-gate circuit, length denotes random operands with the observed gate sequence, and uniform denotes a uniformly random stabilizer state. Both random references exclude exact target matches. Cap is the percentage of all outputs at that qubit count excluded by the 2,000-character limit, not a percentage of $N$. Panel (a) reports mean deficits, with standard errors (SE) across observed outputs in parentheses. Panel (b) gives actual counts for observed and empty candidates and expected counts for the random references.}
\label{tab:32b-reference-strata}
\small
\begin{tabular*}{\columnwidth}{@{\extracolsep{\fill}}rrrcrrr@{}}
\toprule
\multicolumn{7}{l}{(a) Mean constraint deficit} \\
$n$ & $N$ & Cap (\%) & Observed (SE) & Empty & Length & Uniform \\
\midrule
2 & 19 & 0.0 & 1.47 (0.12) & 1.84 & 1.58 & 1.75 \\
3 & 46 & 0.0 & 2.09 (0.12) & 2.24 & 2.51 & 2.65 \\
4 & 95 & 0.0 & 3.11 (0.10) & 3.29 & 3.51 & 3.61 \\
5 & 131 & 0.0 & 3.79 (0.11) & 4.08 & 4.39 & 4.59 \\
6 & 207 & 0.0 & 4.68 (0.10) & 4.84 & 5.37 & 5.58 \\
7 & 279 & 0.0 & 5.68 (0.10) & 5.82 & 6.33 & 6.57 \\
8 & 344 & 0.0 & 6.34 (0.11) & 6.57 & 7.20 & 7.57 \\
\midrule
9 & 437 & 2.3 & 7.09 (0.11) & 7.42 & 8.18 & 8.57 \\
10 & 376 & 33.7 & 7.24 (0.14) & 7.57 & 8.90 & 9.57 \\
11 & 281 & 53.3 & 6.93 (0.17) & 7.68 & 9.52 & 10.57 \\
12 & 366 & 53.4 & 7.46 (0.18) & 8.07 & 10.26 & 11.57 \\
\bottomrule
\end{tabular*}
\par\medskip
\begin{minipage}{\columnwidth}
\small
\begin{tabular*}{\columnwidth}{@{\extracolsep{\fill}}rrrrrr@{}}
\toprule
\multicolumn{6}{l}{(b) Counts with constraint deficit at most two} \\
$n$ & $N$ & Observed & Empty & Length & Uniform \\
\midrule
2 & 19 & 19 & 19 & 19.00 & 19.000 \\
3 & 46 & 30 & 28 & 18.52 & 14.623 \\
4 & 95 & 24 & 17 & 8.06 & 4.178 \\
5 & 131 & 25 & 13 & 3.29 & 0.377 \\
6 & 207 & 28 & 10 & 2.08 & 0.019 \\
7 & 279 & 23 & 8 & 1.39 & $<0.001$ \\
8 & 344 & 36 & 8 & 2.08 & $<0.001$ \\
\midrule
9 & 437 & 34 & 6 & 0.95 & $<0.001$ \\
10 & 376 & 37 & 10 & 1.68 & $<0.001$ \\
11 & 281 & 32 & 9 & 1.80 & $<0.001$ \\
12 & 366 & 46 & 7 & 1.90 & $<0.001$ \\
\bottomrule
\end{tabular*}
\par\smallskip
\parbox{\columnwidth}{\footnotesize Length-reference means average the per-output reference means, and expected counts sum the per-output probabilities. The standard errors in panel (a) differ from the count standard deviations in Table~\ref{tab:32b-reference-comparisons}.}
\end{minipage}
\end{table}

\subsection{Complete Error Distribution}

Table~\ref{tab:32b-error-distribution} gives every constraint-deficit count for the 2,581 valid but state-incorrect outputs in this analysis.
Let $F$ denote squared overlap between the candidate and target states, and let $r$ denote the number of independent signed target constraints missing from their shared stabilizer subgroup.
The table separates zero and nonzero overlap at each deficit and records the observed fidelity value for the nonzero branch.
All counts and overlap summaries in this subsection use the 2,581 errors, excluding the 239 exact outputs and 17 invalid outputs in the larger 2,837-output analysis set.

\begin{table}[htbp]
\centering
\small
\setlength{\tabcolsep}{5pt}
\caption{Complete constraint-deficit distribution of the 2,581 valid, state-incorrect 32B RFT-v1 outputs with saved QASM shorter than 2,000 characters, pooled over two to twelve qubits. Deficit $r$ counts missing independent signed target constraints and $F$ is squared overlap with the target state. The second column gives the fidelity value only for the nonzero-overlap branch. The two count columns partition each row, and none of these outputs prepares the target exactly.}
\label{tab:32b-error-distribution}
\begin{tabular*}{\columnwidth}{@{\extracolsep{\fill}}ccrrr@{}}
\toprule
Deficit $r$ & Nonzero $F$ & $F=0$ count & $F>0$ count & Total \\
\midrule
1 & $2^{-1}$ & 70 & 72 & 142 \\
2 & $2^{-2}$ & 100 & 92 & 192 \\
3 & $2^{-3}$ & 106 & 72 & 178 \\
4 & $2^{-4}$ & 143 & 92 & 235 \\
5 & $2^{-5}$ & 116 & 88 & 204 \\
6 & $2^{-6}$ & 153 & 147 & 300 \\
7 & $2^{-7}$ & 177 & 160 & 337 \\
8 & $2^{-8}$ & 179 & 201 & 380 \\
9 & $2^{-9}$ & 157 & 178 & 335 \\
10 & $2^{-10}$ & 82 & 95 & 177 \\
11 & $2^{-11}$ & 37 & 34 & 71 \\
12 & $2^{-12}$ & 15 & 15 & 30 \\
\midrule
All & & 1,335 & 1,246 & 2,581 \\
\bottomrule
\end{tabular*}\par\vspace{-0.75pt}
\end{table}

Nonzero overlap occurs in 1,246 of these 2,581 errors, or 48.28\%.
Across the full error subset, mean squared overlap is 0.03150, the median is 0, the 90th percentile is 0.0625, and the maximum is 0.5.
The median constraint deficit is 7.
Among the 1,335 zero-fidelity outputs, the median deficit is 6 and 170 outputs (12.7\%) lack at most two independent constraints.
Thus zero fidelity can coexist with few missing constraints, although most outputs in this subgroup differ in more than two independent constraints.
The counts are pooled over two to twelve qubits, with the corresponding size-specific reference comparisons in Table~\ref{tab:32b-reference-strata}.

\section{Failure Taxonomy}
\label{app:failure-taxonomy}

We group failures into categories observable from verifier outputs.
Table~\ref{tab:failure-breakdown} reports the direct decomposition, while Table~\ref{tab:appendix-search-outcomes} reports target-level search coverage.
These categorical outcomes are distinct from the state-overlap and constraint-deficit analysis in Appendix~\ref{app:32b-reference-comparisons}.

\noindent\textbf{Strict saved-label mismatches dominate after code-level validation.}
Table~\ref{tab:failure-breakdown} decomposes the 32B direct-verifier outcomes into syntax failures, non-Clifford valid parses, exact saved-label matches, and valid Clifford circuits with a saved-label mismatch.
Across all three rows, syntax failures are rare and non-Clifford valid parses do not appear, while saved-label mismatches account for more than 96\% of held-out prompts.
The table reports the strict categories, while Appendix~\ref{app:32b-reference-comparisons} uses StateEq, state overlap, and shared stabilizer constraints to distinguish representation mismatch from state discrepancy among the RFT-v1 errors.
The result shows that passing the code-level checks is distinct from satisfying the strict target criterion.

\begin{table}[htbp]
\centering
\scriptsize
\setlength{\tabcolsep}{2pt}
\caption{32B direct-verifier outcomes over 3,894 held-out prompts. Counts use the strict saved-label decision and separate exact matches from syntax failures, non-Clifford parses, and valid Clifford outputs with a saved-label mismatch.}
\label{tab:failure-breakdown}
\begin{tabular*}{\columnwidth}{@{\extracolsep{\fill}}lrrrr@{}}
\toprule
System & Exact & Syntax fail & Non-Cliff. & Label mismatch \\
\midrule
Basic AG-CoT & 102 & 5 & 0 & 3787 \\
GRPO-v2 & 86 & 5 & 0 & 3803 \\
RFT-v1 & 122 & 17 & 0 & 3755 \\
\bottomrule
\end{tabular*}\par\vspace{-0.5pt}
\end{table}
\FloatBarrier

\subsection{Depth Sensitivity}

For GRPO-v2, syntax and Clifford validity are nearly saturated, but exact target matching varies substantially with the number of layers in the reference circuit.
Table~\ref{tab:depth-breakdown} groups the GRPO-v2 direct evaluation by this layer count.
Exact@1 is 74.5\% for the 47 targets with one or two layers and falls to 0.7\% for the 142 targets with 15 to 20 layers.
There are no strict target matches among the 664 targets with 21 to 50 layers and one among the 2,743 targets with more than 50 layers.
These counts show that correct outputs are concentrated among shallow targets, while exact target matching is rare in the deeper groups that contain most test examples.

\begin{table}[H]
\vspace{-0.75pt}
\centering
\footnotesize
\setlength{\tabcolsep}{2pt}
\caption{GRPO-v2 direct evaluation by target circuit-layer bucket under strict saved-label Exact@1. Bucket percentages are rounded to one decimal place. The last bucket contains one correct output, whose rate of approximately 0.0365\% rounds to 0.0\%.}
\label{tab:depth-breakdown}
\begin{tabular*}{\columnwidth}{@{\extracolsep{\fill}}lrrr@{}}
\toprule
Layer bucket & Correct / total & Exact@1 (\%) & Test share (\%) \\
\midrule
1--2 & 35 / 47 & 74.5 & 1.2 \\
3--5 & 27 / 64 & 42.2 & 1.6 \\
6--10 & 17 / 132 & 12.9 & 3.4 \\
11--14 & 5 / 102 & 4.9 & 2.6 \\
15--20 & 1 / 142 & 0.7 & 3.6 \\
21--50 & 0 / 664 & 0.0 & 17.1 \\
51+ & 1 / 2743 & 0.0 & 70.4 \\
\midrule
All & 86 / 3894 & 2.21 & 100.0 \\
\bottomrule
\end{tabular*}\par\vspace{-13.25pt}
\end{table}

\subsection{Uncertainty and Search Outcomes}

Tables~\ref{tab:stateeq-audit} and~\ref{tab:appendix-exact-bootstrap-ci} report uncertainty for state equivalence and strict saved-label matching, respectively.

\begin{table}[H]
\vspace{-1pt}
\centering
\scriptsize
\setlength{\tabcolsep}{2pt}
\caption{State-equivalence confidence intervals over 3,894 held-out prompts; \textit{lab.} denotes the saved-label verifier convention.}
\label{tab:stateeq-audit}
\begin{tabular*}{\columnwidth}{@{\extracolsep{\fill}}lccc@{}}
\toprule
& \multicolumn{2}{c}{Direct target match} & \\
\cmidrule(lr){2-3}
System & Exact@1$_{\mathrm{lab.}}$ (\%) & StateEq@1 (\%) & 95\% CI \\
\midrule
Zero-shot & 0.10 & 0.33 & [0.15, 0.52] \\
Basic AG-CoT & 2.62 & 5.37 & [4.67, 6.06] \\
Step AG-CoT & 2.31 & 4.83 & [4.15, 5.50] \\
GRPO-v2 & 2.21 & 5.26 & [4.57, 5.98] \\
RFT-v1 & \best{3.13} & \best{6.14} & [5.37, 6.91] \\
\bottomrule
\end{tabular*}\par\vspace{-5pt}
\end{table}

\begin{table}[H]
\centering
\footnotesize
\setlength{\tabcolsep}{4pt}
\caption{Bootstrap confidence intervals for the direct Exact@1 metric (\%) computed over 10,000 resamples of the 3,894 held-out prompts.}
\label{tab:appendix-exact-bootstrap-ci}
\begin{tabular*}{\columnwidth}{@{\extracolsep{\fill}}lrrc@{}}
\toprule
Run & Exact successes & Exact@1 (\%) & 95\% CI \\
\midrule
Basic AG-CoT & 102 / 3894 & 2.62 & [2.13, 3.13] \\
GRPO-v2 & 86 / 3894 & 2.21 & [1.77, 2.70] \\
RFT-v1 & 122 / 3894 & 3.13 & [2.59, 3.70] \\
\bottomrule
\end{tabular*}\par\vspace{-5pt}
\end{table}

\begin{table}[H]
\centering
\footnotesize
\setlength{\tabcolsep}{2.5pt}
\caption{32B strict saved-label search coverage over 3,894 held-out prompts.}
\label{tab:appendix-search-outcomes}
\begin{tabular*}{0.82\columnwidth}{@{\extracolsep{\fill}}lrr@{}}
\toprule
Search run & Covered & Missed \\
\midrule
RFT-v1 Best-of-64 & 202 & 3692 \\
\bottomrule
\end{tabular*}\par\vspace{-9pt}
\end{table}
Search miss means that none of the 64 sampled candidates for a held-out prompt passed exact verification.
The direct failure categories and their interpretation are given with Table~\ref{tab:failure-breakdown}.

\end{document}